\documentclass{article}

 \usepackage[preprint]{neurips_2026}

\usepackage[utf8]{inputenc} % allow utf-8 input
\usepackage[T1]{fontenc}    % use 8-bit T1 fonts
\usepackage{hyperref}       % hyperlinks
\usepackage{url}            % simple URL typesetting
\usepackage{booktabs}       % professional-quality tables
\usepackage{amsfonts}       % blackboard math symbols
\usepackage{nicefrac}       % compact symbols for 1/2, etc.
\usepackage{microtype}      % microtypography
\usepackage{xcolor}         % colors\

\usepackage{graphicx}
\usepackage{wrapfig}
\usepackage{subcaption}
\usepackage{tabularx}
\usepackage{listings}
\usepackage{amsmath}
\usepackage{cleveref}
\usepackage{enumitem}
\usepackage{wasysym}
\usepackage{tcolorbox}
\tcbuselibrary{breakable}
\usepackage{caption}

\usepackage{mdframed}

\setcitestyle{round}

\definecolor{shadecolor}{HTML}{F4F4F4}

\lstdefinestyle{prompt}{
  backgroundcolor=\color{shadecolor},
  basicstyle=\ttfamily\footnotesize,
  breaklines=true,
  frame=single,
  framerule=0.4pt,
  rulecolor=\color{gray!50},
  xleftmargin=8pt,
  xrightmargin=8pt,
  aboveskip=4pt,
  belowskip=4pt,
}

\title{SLAyiNG: A Diverse and Community-validated Dataset of Queer Slang}

\author{%
  Leonor Veloso\textsuperscript{1,2} \quad Lea Hirlimann\textsuperscript{1,2} \quad
  Lucija Mihić Zidar\textsuperscript{3} \quad
  Philipp Wicke\textsuperscript{4}
  \\
  \textbf{Valentin Hofmann}\textsuperscript{1,2} \quad \textbf{Hinrich Schütze}\textsuperscript{1,2}
  \\
  \textsuperscript{1}Center for Information and Language Processing, LMU Munich \\
  \textsuperscript{2}Munich Center for Machine Learning (MCML) \\
  \textsuperscript{3}Brandenburg Technical University Cottbus–Senftenberg \qquad \textsuperscript{4}Auryal GmbH \\
  \texttt{lveloso@cis.lmu.de}
}

\begin{document}

\maketitle

\begin{abstract}

Queer vernacular is rarely studied in NLP, despite advancements in resources and evaluation for other sociolects and informal language. Because of this, NLP systems often process queer language incorrectly, e.g., they misclassify it as hate speech or generate negative responses. To address this problem, we propose \textsc{Slaying}, the first real-world dataset of English queer slang. \textsc{Slaying} is community-validated,  and includes over 500 queer slang terms that pertain to more than 20 queer subcommunities. We argue that queer language data resources have great potential in NLP -- e.g., as components of large pretraining corpora and as the basis for benchmarks -- and can improve queer users' experience of NLP systems. We leverage \textsc{Slaying} for two novel findings in support of this argument: (i) For a number of language models, we show that they are unbiased towards the queer community, but at the same time unable to process its language, i.e., absence of representation bias does not entail the absence of linguistic bias. (ii) Model performance on queer slang varies across queer subcommunities; it is generally worse for slang pertaining to African-American and Latine communities. These findings are relevant for both the queer NLP and the broader ML communities. \textsc{Slaying} is available to the public, and open to future revisions and extensions. \textbf{Warning:} This paper contains profane and potentially offensive language.
\end{abstract}

\section{Introduction}
Language is an \textit{act of identity}, a tool for self-identification and recognition within one's community \citep{foucault2019history, cameron2006language}. This aspect of language is especially meaningful for queer individuals, for whom queer slang can signify cultural positioning, promote social cohesion within the community, and serve as form of protest, evolving along socio-political changes \citep{kaabour2025introducing}. Queer slang also enriches language -- as \citet{rodgers1972queens} puts it, ``\textit{Camp}, \textit{straight}, and \textit{vibrations} were once solely gay inventions that are today part of every hip vocabulary''.

Regardless of extensive research efforts in the fields of queer linguistics \citep{kulick2000gay} and information activism targeted at archiving the collective language of the queer community \citep{cifor2023mediating}, queer slang (or sociolect) is not adequately represented in evaluation benchmarks for NLP systems -- the scope of current literature is either limited to a single feature of queer language, such as reclaimed slurs \citep{dorn2024harmful}, or to small lists of queer slang terminology used in synthetic prompts \citep{tint2025guardrails, basoah2025not}. In short, we have a limited understanding of how linguistics biases in NLP and AI systems may worsen the experiences of queer stakeholders -- which are already underserved by these systems through misgendering \citep{hossain2023misgendered}, stereotyping \citep{felkner2023winoqueer}, and lower performance in tasks such as speech recognition \citep{sanchez2024beyond} and coreference resolution \citep{cao2020toward, gautam2024robust}. As progress is made towards improved language processing of other English sociolects, such as AAVE \citep{hofmann2024ai, gupta2024aavenue, nias2025culture}, and informal language \citep{sun2024toward}, it is timely to improve resources and evaluation of queer vernacular. A sound evaluation of anti-queer linguistic biases requires diverse and ecologically valid resources that include the language of many subcommunities, are non-synthetic, and loop in the queer community during its curation process; yet such resources are currently lacking. Here, we fill this gap. Specifically, we make the following three contributions:

(i) We propose \textsc{Slaying}\footnote{Available at \url{https://github.com/leonorv/slaying-dataset}. The title is meant as a pun, as \textit{slaying} includes \textit{slay} + \textit{slang}.}, the first real-world dataset of queer slang, containing more than 500 terms, used in a sentence-wise context, amounting to 3405 examples. \textbf{\textsc{Slaying} is non-synthetic, manually annotated, contains slang pertaining to diverse queer subcommunities, and is community-validated -- these features contribute to its ecological validity and differentiate it from existing resources.} \textsc{Slaying}'s vocabulary, sentences, and metadata are released under CC by 4.0 where legally permissible. Our extensive data curation process offers insights on inter-annotator agreement and variation, while ensuring the final public data is correct and safe. The non-synthetic and diverse nature of \textsc{Slaying} makes it a good candidate for perplexity analysis -- following prior work on other English sociolects and dialects \citep{lin2025assessing, tan2020mind} -- and toxicity analysis -- following prior work on features of queer vernacular \citep{dorn2024harmful}. We use these methods in the case studies described below.

(ii) We showcase \textsc{Slaying} as a vehicle to measure linguistic bias in language models, discussing its relationship with representational harms against the queer community. Evaluating \textsc{DataDecide} models \citep{magnussondatadecide} across pretraining data mixes, we observe a negative correlation between perplexity on \textsc{Slaying} and representation bias against queer commmunities, measured on \textsc{WinoQueer}. \textbf{This suggests that bias in LLMs is not a monolith -- a model can be unbiased towards a community while performing poorly on that community's language.} In our experiments, we show that this can be overcome by exposing language models to harmless, ingroup examples of queer language -- continued pretraining an OLMo-1B model on \textsc{Slaying} successfully reduces its bias, as tested on \textsc{WinoQueer}. 

(iii) We showcase \textsc{Slaying} as a resource to perform fine-grained and intersectional analyses of the slang of various queer subcommunities. These analyses are unlocked throught the diverse nature of \textsc{Slaying}, whose slang terms pertain to more than 20 queer subcommunities, a substantially larger number than in prior datasets. We find that these are not served equally -- models perform worse in text containing non-binary and drag slang. \textbf{Importantly, models also perform worse on slang of African-American and Latine communities, compared to slang terms not specified for ethnicity.} Terms pertaining to these communities are also classified as more toxic by hate speech detection systems, which effectively contributes towards the censoring of already marginalized groups. % motivate 

% IN PROGRESS
 \begin{figure}
   \centering
   \includegraphics[scale=0.6]{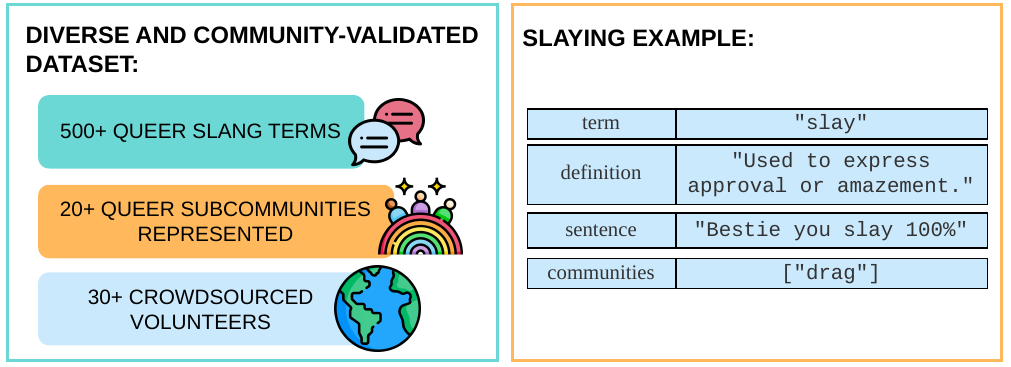}
   \caption{Overview of \textsc{Slaying}.}
   \label{fig:slaying_overview}
 \end{figure}

\section{Background and Related Work}
\label{sec:Background and Related Work}

Glossaries of terms pertaining to the LGBTQ+ community were a prominent research object during the emergence of queer linguistics \citep{livia1997queerly}. While the concerns of that field have shifted and broadened since the end of the 20th century \citep{leap2015queer, zimman2018transgender}, efforts towards the archival of queer language continue -- and are of relevance to medicine \citep{kronk2020development}, preservation of linguistic history \citep{rawson2010archiving}, and online and offline access to reliable queer-related information \citep{rawson2009accessing}. \textsc{Slaying} makes use of these glossaries, linked vocabularies, and ontologies, owing its existence to this body of work.

Queer language as a research object extends to NLP -- the subfield of research that centers queer individuals and communities as stakeholders of NLP systems is sometimes referred to, in academic circles and publications, as Queer NLP \citep{weber2026queer}. Many of these works are focused on resources and evaluation -- for instance, \citet{felkner2023winoqueer} propose \textsc{WinoQueer}, a community-in-the-loop benchmark for queerphobic bias in LLMs. To highlight some notable examples of works that focus on queer slang and vernacular: \citet{dorn2024non} propose \textsc{QueerReclaimLex}, a template-based, dataset of usages of reclaimed queer slurs; \citet{tint2025guardrails} show that prompts containing queer slang elicit more negative responses from LLMs, relative to heteronormative prompts; and \citet{basoah2025not} find that queer participants feel more social presence from an LLM that uses queer slang. Unlike this prior work on NLP and queer vernacular, \textsc{Slaying} is non-synthetic, and contains slang pertaining to different communities and etymological background. To align ourselves with this line of work (and other notable works on informal language in NLP, such as \citet{sun2024toward}), we refer to \textsc{Slaying} as a dataset of \textit{queer slang}.

In this work, we define \textit{queer slang} as the \textit{lingo}, \textit{argot}, or \textit{sociolect} pertaining to gay, lesbian, transgender, or any other LGBTQ+ community \citep{kulick2000gay}. This is not without its caveats, which we discuss in the Limitations section. Note that term \textit{slang} itself is not statically defined within linguistics, due to its nature as a ``deviant'' repertoire of phrases and words \citep{agha2015tropes}. A detailed overview of our related work and background can be found in \Cref{app:Related Work and Background: Details} of the Appendix.

\section{Creating \textsc{Slaying}}

\subsection{Sourcing Queer Slang Terms}
\label{Term Sources}

\paragraph{GSSO and lgbtDB} 

GSSO \citep{kronk2020development} is a manually curated open-source ontology utilizing related glossaries from biology, medicine, psychology, sociology, and gender studies. At the time of writing, GSSO is no longer being updated, and its database is being migrated to lgbtDB\footnote{\url{https://lgbtdb.wikibase.cloud/wiki/Main_Page}}. For our purposes, we scrape all instances from queer slang-related classes of GSSO and lgbtDB (details regarding these classes can be found in \Cref{app: Sources: Details}), yielding a total of 414 (for GSSO) and 215 (for lgbtDB) terms.

\paragraph{Chew Inclusive Terminology Glossary}

The Chew Glossary \citep{chew2023inclusive} contains a section of LGBTQIA+ Slurs and Slang, containing 65 slang terms. The glossary contains two additional lists, one for slurs targeted at gay men, another targeted at women. However, some of the terms in these lists are present in our other sources as ``slang'', as the offensiveness of a term is dependent on its contextual usage -- we rely on manual annotation to ensure the safety of the final dataset.

\paragraph{Wiktionary}

Wiktionary is a web-based, user-curated dictionary of terms that has seen extensive usage in various NLP tasks \citep{zesch2008extracting, navarro2009wiktionary, li2012wiki}. Using the WiktionaryParser\footnote{\url{https://github.com/suyashb95/WiktionaryParser}} tool, we parse through the category of LGBTQ English slang\footnote{\url{https://en.wiktionary.org/wiki/Category:English_LGBTQ_slang}}. This category includes 56 terms (which we refer to as simply ``LGBTQ slang''), and 4 subcategories: ``English 4chan /lgbt/ slang'', ``English drag slang'', ``English gay slang'', and ``English transgender slang''. This yields a total of 251 terms.

All sources were scraped between the dates of 03/07/2025 and 30/07/2025, yielding a total of 695 terms and 90 variants -- which, depending on the source, can be synonyms or alternative spellings of the term (e.g., \textit{faux queen} and \textit{bio queen}, referring to a cis female drag performer). Naturally, our sources often overlap. Details on handling overlapping terms and similar definitions can be found in \Cref{app:Definition Cleaning: Details}.

\subsection{Sourcing Raw Data and Required Annotation Tasks}

We curate a sentence-wise dataset (following the dataset of English slang, curated by \citet{sun2024toward}), each example depicting usage of the terms described in \Cref{Term Sources} -- an unfiltered version of \textsc{Slaying}. It is comprised of examples derived from three sources: movie and TV show subtitles (from the OpenSubtitles Corpus \citep{lison2016opensubtitles2016}), social media posts (from the forum-based social media platform Reddit\footnote{\url{https://www.reddit.com/}}), and podcast transcripts (from the publicly available platform Podscripts\footnote{\url{https://podscripts.co}}), in an attempt to reflect real-world usage across contexts. Details on these sources, including our efforts regarding anonymization, licensing, and preservation of metadata can be found in the Appendix, \Cref{app: Sources: Details,app:Licensing}.

%\subsection{Annotation Tasks}

Our complete raw dataset contains 197,958 examples. To ensure a correct and safe final version of \textsc{Slaying}, filtering and annotation are required for three tasks, which we justify and describe below:

\paragraph{Sense Disambiguation}  Several terms of the slang term list have common, not-queer related senses (see \textit{mother} or \textit{read}). This entails that the biggest and most consuming annotation task is sense disambiguation, since the majority of the examples from our raw dataset are false positives. As such, during the pre-filtering stages of our data curation and annotation pipeline, we only annotate for this task. Some terms have several distinct definitions in the context of queer slang. For those cases, we consider all valid definitions during annotation.
    
\paragraph{Harmful speech detection} Approximately 15\% of our terms are described in their original sources as slurs, derogatory, or potentially offensive. This makes it likely that an example found in our raw data is harmful. These terms appear in our cited slang sources due to processes of reclamation \citep{popa2020reclamation} and/or ingroup usage. Notably, some of these terms have appeared and been considered ``queer slang'' (in their reclaimed sense) in other NLP works \citep{tint2025guardrails}, such as \textit{fag} and \textit{homo}. The inclusion of these repurposed terms in data is important in the context of hate-speech detection, since classifiers that falsely identify harm run the risk of suppressing speech \citep{dorn2024harmful,anderson2013did}. We recognize that whether a particular use of a term is considered offensive or not depends on the term's target group, and even then, it is unclear how many members of a group must declare its offensiveness for the usage to be considered harmful. 
    
\paragraph{Author identity identification} We base our definition of \textit{ingroup} and \textit{outgroup}, as well as the setup for the annotation for this task, on the work of \citet{dorn2024harmful}. In a sentence with an identity term or slur, we say the ingroup is the population referenced by the identity term or slur's neutral correlate (e.g., the neutral correlate for \textit{dyke} is \textit{lesbian}). The outgroup is the population not referenced by the identity term or neutral correlate. Author ingroup membership has an impact on the intended and perceived harm of a sentence. \citet{dorn2024harmful} find that ingroup usages of slurs are much less likely to be considered harmful. As such, this annotation task serves as additional context for filtering and annotating potentially harmful sentences.

\subsection{Pre-filtering}
\label{subsec:Pre-filtering}

Due to the size of our raw dataset (197,958 examples), manually annotating every example would be unfeasible. As we expect the majority of the raw dataset to be false positives for the task of sense disambiguation (due to the ambiguous nature of many queer slang terms), we evaluate the potential of LLMs as pre-filtering models. To create a gold standard for the task of sense disambiguation, three of the authors start by annotating 25 examples -- for this , we adapt the task design of \citet{erk2013measuring}, where the applicability of each dictionary sense (our term definitions) is rated on a Likert scale from 1 (not at all) to 5 (completely). This initial effort yields a Krippendorff's $\alpha$ of $0.89$. To ensure full coverage of terms in this experiment, we created an additional data subset consisting of up to 15 examples per term. Because of their high internal agreement, these were split across the three author annotators. From this subset, 250 examples \footnote{Due to computing and budgeting restraints.} were used to test a small suite of OpenAI models: o4-mini with and without high reasoning, and GPT-5 Mini with high reasoning. These achieve micro F1 scores of $0.78$, $0.77$, and $0.79$, respectively. GPT-5 Mini also achieves the lowest ratio of low-agreement terms (whose instances are scored under $0.7$ F1), at $22.8\%$. Choosing GPT-5 Mini as our best model, we evaluate it on the full gold standard set, yielding an F1 score of 0.78, and let it label all examples. Further details and discussion on LLMs as pre-filtering models in the context of queer slang can be found in \Cref{app:Prefiltering: Details}.

\subsection{Internal Annotation}
\label{subsec:Internal Annotation}

Our internal annotation process follows the principles of agile corpus creation, following the description and recommendations of \citet{klie2024analyzing}. Following the pre-filtering step, we choose 5973 examples\footnote{The hypothesis being that, even if half of these examples were discarded, the final dataset size would still be comparable to other slang datasets (+2k manually annotated examples, from \citet{sun2024toward})} that were labeled by the model as correct, while attempting to have a balanced and non-repetitive sample to annotate. 

The first author (which serves as the principal annotator, henceforth PA) designed the annotation guidelines. These guidelines were used in a pilot study, where the PA and three other authors (henceforth IAs, or internal annotators) label a sample of 25 examples. This is a different annotation effort from the one described in \Cref{subsec:Pre-filtering}, which solely focused on sense disambiguation. Here, annotators label examples for sense disambiguation (``Does the given definition apply to the term as used in the sentence'', which can be labeled as \texttt{Yes}, \texttt{No}, or \texttt{I don't know}), harmful speech detection (``Is the given sentence harmful?'', which can be labeled as \texttt{Yes}, \texttt{No}, or \texttt{I don't know}), and author identity identification (``Is the author of the sentence an ingroup member'', which can be labeled as \texttt{Yes}, \texttt{Implied}, \texttt{Unclear}, \texttt{No}, or \texttt{The sentence does not contain an identity term}). This effort corresponded to an average pairwise percent agreement of 0.75 (sense disambiguation), 0.78 (harmful speech detection), and 0.53 (author identity identification). After discussion of disagreements and refinement of the guidelines, the agreement increased to 0.81, 0.88, and 0.75. 

The 5973 selected examples were divided in 3 \textit{PA batches} (annotated solely by the PA) and 2 \textit{SC batches} (sanity check batches of 100 examples, annotated by the PA and IAs). After completion of a PA batch, an SC batch was annotated to evaluate stability of agreement and offer opportunities to further refine the guidelines. The labels for SC batch examples are decided through a majority vote (if there is no majority label, the example is discarded). \emph{The manual internal annotation process yielded our final version of \textsc{Slaying}, consisting of 3405 entries annotated as harmless} (labeled as ``not harmful'' for the task of harmful speech detection). In alignment with our goal of not censoring the speech of the queer communities, we also release a private subset of \textsc{Slaying}, containing further 121 examples that were labeled as uncertainly harmful (\texttt{I don't know} for the task of harmful speech detection), but were considered ingroup speech (\texttt{Yes} or \texttt{Implied} for the task of author identity identification). This dataset is available only upon direct contact with the authors, due to its higher potential for misuse. The final annotation guidelines used and refined during this process can be found in Appendix \Cref{app: Annotation Sheet and Guidelines}.

\subsection{Crowdsourced Annotation}
\label{subsec:Crowdsourced Annotation}

% \begin{figure}
%   \centering
%   \includegraphics[scale=0.45]{images/public_slaying_agreement.pdf}
%   \caption{Inter-annotator agreement between internal and crowdsourced annotators, for the publicly available version of \textsc{Slaying}. Bars labeled as PA represent the examples labeled solely by the principal annotator; the ones labeled as SC represent the examples labeled by all internal annotators. \textbf{Agreement is moderate to high for sense disambiguation (Task 1) and harmful speech detection (Task 2), validating the correctness and safety of our data.}}
%   \label{fig:crowdsouced_agreement_public}
% \end{figure}

%validate corpus (one example per term), split into 22 partitions of each 20 terms --> 440 samples, favoring samples of the private subset of Slaying. 31 annotators, timeframe x, what demographics, screener, vis in the appendix. report on interannotator agreement, task wise agreement

%The manual internal annotation process yielded our final version of \textsc{Slaying}, consisting of TODO entries. 
To ensure the validity of our data and directly involve the community in the curation of \textsc{Slaying}, we conducted a crowdsourced annotation study with 33 voluntary participants. Each participant was asked to label 20 examples of \textsc{Slaying}, following the same task design and annotation guidelines as internal annotation (\Cref{subsec:Internal Annotation}). The pool of participants was diverse, containing cis and trans participants of multiple gender identities and sexualities (\Cref{App:Demographics}).

A total amount of 12\% of \textsc{Slaying}, split into 22 partitions, was selected for crowdsourced annotation by sampling one usage example per slang term. This sampling is purposely biased towards the private subset of \textsc{Slaying} to ensure coverage of sensitive cases. As a result of the sampling bias, 86 sentences were sampled from the private subset, constituting 19.91\% of the crowdsourced annotated data and covering more than 70\% of the private dataset. Overall, half of the partitions were double-annotated to enable agreement analysis, while the remaining partitions were single-annotated. Complimentary to the annotation, we collected demographic information including age, gender, sexuality, trans identity, English proficiency, as well as self-reported familiarity and usage of queer slang (questions were retrieved from \citet{basoah2025not}). Agreement between crowdsourced annotators and the principal annotator is moderate to high for sense disambiguation and harmful speech detection ($\text{Gwet's AC}_1 = 0.79 \pm 0.16$ and $0.66 \pm 0.24$, respectively). We extrapolate these results to the remainder of \textsc{Slaying}, and argue in favor of its correctness and safety. Agreement is low for the task of author identity identification ($\text{Gwet's AC}_1 = 0.22 \pm 0.22$), which we attribute to the high subjectivity of the task. However, labels for this task only affect the private version of \textsc{Slaying}, as all examples that were labeled as correct and harmless enter its public version. Full details are  provided in \Cref{App:Demographics,app:Additional Questionnaire for Crowdsourced Annotators,app:Inter-annotator Agreement}.

%talk about agreement with PA/SC, agreement private vs public, inter annotator, volunteers with both low English proficiency and no experience with English Queer Slang were excluded from the Evaluations

%\begin{figure}
%    \centering
%    \includegraphics[scale=0.45]{images/principal_crowdsourced_matrix.pdf}
%    \caption{Label distribution of \textsc{Slaying examples} annotated by the %principal annotated and crowdsourced annotators. 86\% of examples labeled %as correct by the PA are similarly labeled as correct by the crowsourced %annotators -- validating the correctness of the dataset. Only 6\% of the %examples labeled as harmless by the principal annotator are labeled as %harmful by the crowdsourced annotators, validating the safety of the %dataset.}
%    \label{fig:principal_crowdsourced_main}
%\end{figure}

\section{Experimental Setup}
\label{sec:Experimental Setup}

We present two case studies of analyses that are made possible with the public version of \textsc{Slaying}: \Cref{subsec: Case Study I} discusses the relationship between representation and linguistic anti-queer bias; \Cref{subsec: Case Study II} is a fine-grained analysis of language model performance in specific subcommunities and terminology present in \textsc{Slaying}, through an intersectional lens.

\subsection{Case Study I: Representational Harm and Linguistic Bias}
\label{subsec: Case Study I}

\begin{figure}[]
    \centering
    \includegraphics[scale=0.25]{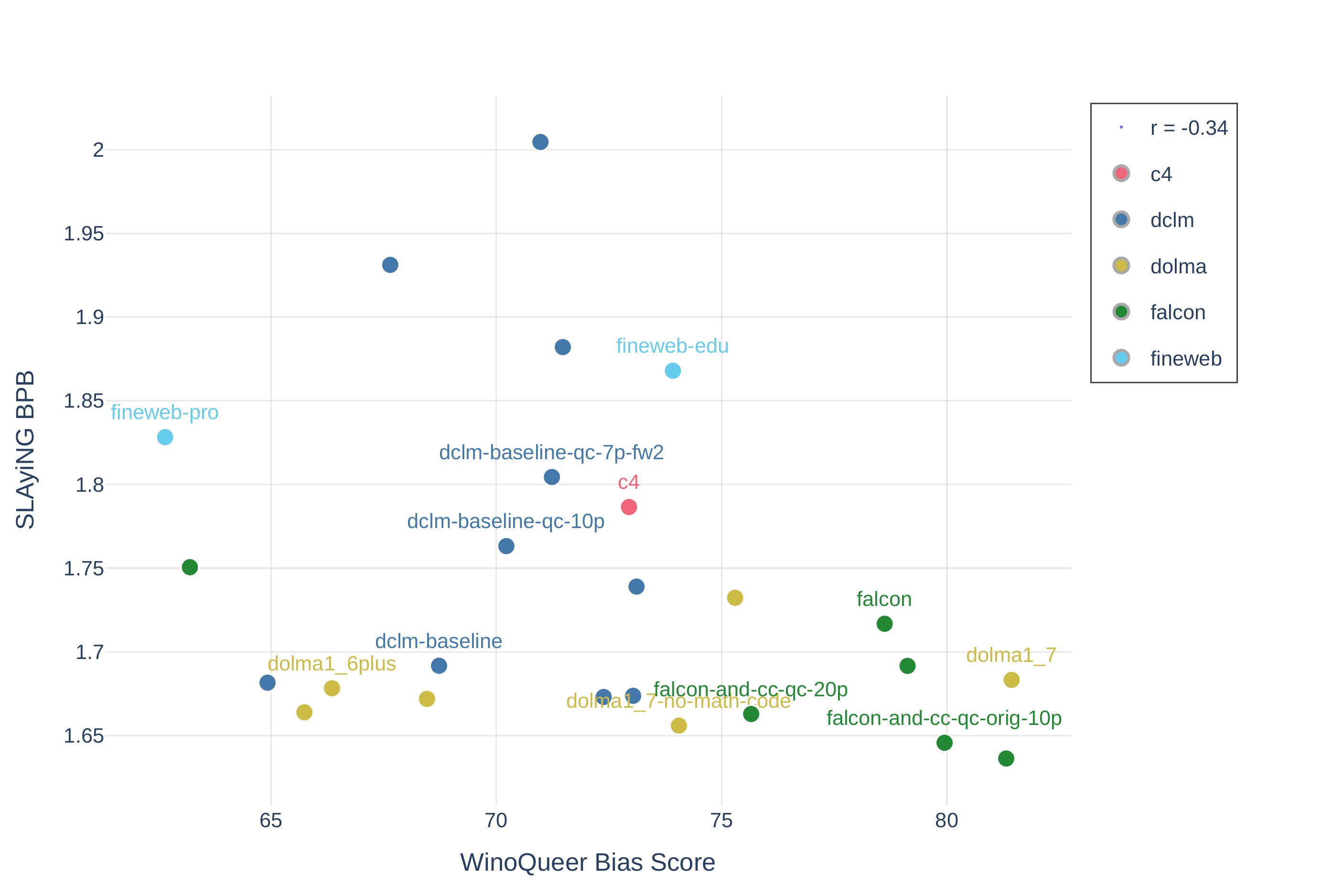}
    \caption{Average BPB of \textsc{Slaying} and \textsc{WinoQueer} bias score for each data recipe of the \textsc{DataDecide} 1B model suite. We find a surprising negative correlation: unbiased models (low bias score) have a lower performance on queer language modeling (high BPB). \textbf{A model can be unbiased towards a community while performing poorly on that community's language.}}
    \label{fig:datadecide_winoqueer}
\end{figure}

% rep harms vs ling bias
% ling bias -> ppl is used to analyze it
% interest to broad ml community -> datadecide/importance of pretraining corpora in learned model biases

\begin{wrapfigure}{r}{0.45\textwidth}
  \centering
  %\vspace{-10pt}  % optional: reduce top whitespace
  \includegraphics[scale=0.4]{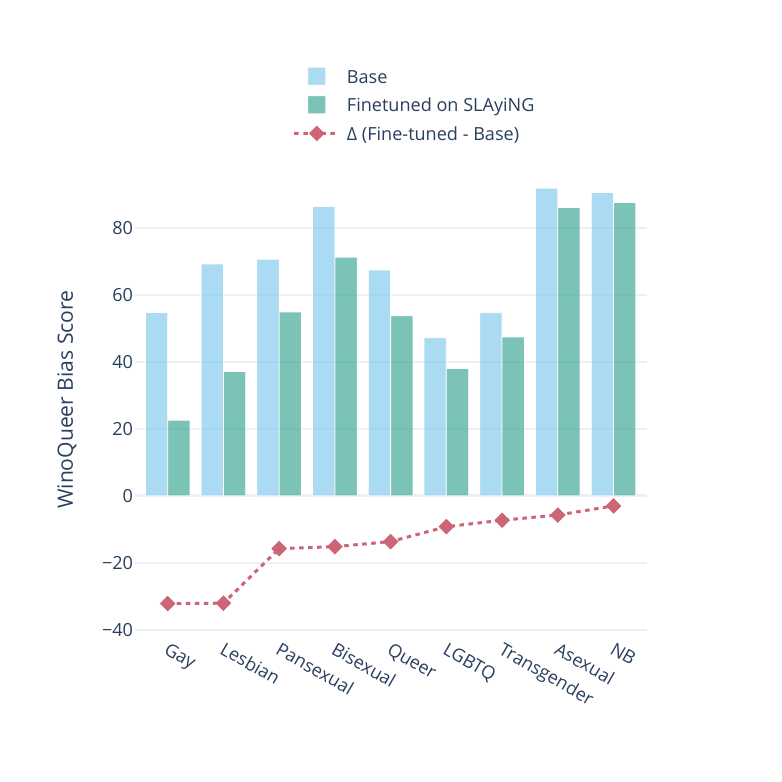}
  \caption{\textbf{Positive effects of fine-tuning an Olmo model on \textsc{Slaying}.} The WinoQueer bias score decreases relative to the base model, for all communities.}
  \label{fig:olmo_improvement}
  %\vspace{-10pt}  % optional: reduce bottom whitespace
\end{wrapfigure}

Anti-queer (or queerphobic) bias in language modeling has so far been mostly studied under the lens of representational harms -- i.e., how language models stereotype queer identities or how system performance differs for queer or cisheterosexual stakeholders. The breadth in vocabulary and non-synthetic nature of \textsc{Slaying} provides a rich testbed to measure linguistic bias, which allows us to perform the first comparison of anti-queer representation and linguistic bias. For this purpose, we employ perplexity as a proxy for language model fit and general performance on text containing queer slang, as is the case for works on other non-standard English variations, such as AAVE \citep{lin2025assessing, garba2024transformer, deas2023evaluation}. Following recommendations from \textsc{Paloma} \citep{magnusson2024paloma} and \textsc{The Pile} \citep{gao2020pile}, we use bits per byte (BPB) as a language fit measurement, evaluated on individual documents rather than on concatenated documents satisfying a maximum sequence length. In order to link our goal with the interests of the broader machine learning community, for this case study, we employ the \textsc{DataDecide} suite \citep{magnussondatadecide}, which includes model sizes up to 1B, using 25 corpora with difference sources (or ``data recipes''). This allows us to analyze which data mixes provide better fit to queer vernacular.

To assess representation bias, we employ the only (at time of writing) community-informed benchmark for anti-queer bias in language models, \textsc{WinoQueer} \citep{felkner2023winoqueer}. \textsc{WinoQueer} is an anti-LGBTQ+ bias benchmark informed by the expressed needs and concerns of the LGBTQ+ community. \textsc{WinoQueer} allows us to evaluate whether our findings regarding language model fit and toxicity classification extrapolate to queerphobic bias benchmarks.  \Cref{fig:datadecide_winoqueer} depicts the average BPB of \textsc{Slaying} (lower is better) and \textsc{WinoQueer} bias score (lower is better, indicating less biased behavior) for each data recipe of \textsc{DataDecide} 1B models. Surprisingly, we find a \emph{negative} Spearman's correlation coefficient ($-0.34$, 95\% CI $[-0.67, 0.07]$, $p = 0.09$) between these metrics. \emph{This suggests that representation and linguistic bias may be partially decoupled}.

We hypothesize that this result is due to the content of the data recipes: if a corpus contains instances of queer slang or queer-related content (thus leading to a lower BPB on \textsc{Slaying}), it is likely that those instances reinforce queerphobic stereotypes, either by being actively harmful, or simply by being authored by outgroup members and thus not representing the experiences of the queer community. To validate this hypothesis, we produce a version of OLMo-1B (as the \textsc{DataDecide} suite uses OLMo's model ladder configurations), further pretrained on \textsc{Slaying}. This model exhibits a lower WinoBias score for all community subsets, as well as having a lower average BPB on \textsc{Slaying} ($\Delta-0.27$). Results are depicted on \Cref{fig:olmo_improvement}, showing that \emph{exposing language models to ingroup, harmless examples of queer language has the potential to both mitigate representational and linguistic bias}. Details, including hyperparameters for the latter model, can be found in \Cref{app:Details of Case Study I}.

\subsection{Case Study II: Community-specific and Term-specific Biases}
\label{subsec: Case Study II}

\begin{figure}
  \centering
  \begin{subfigure}{0.47\textwidth}
    \centering
    \includegraphics[width=\textwidth]{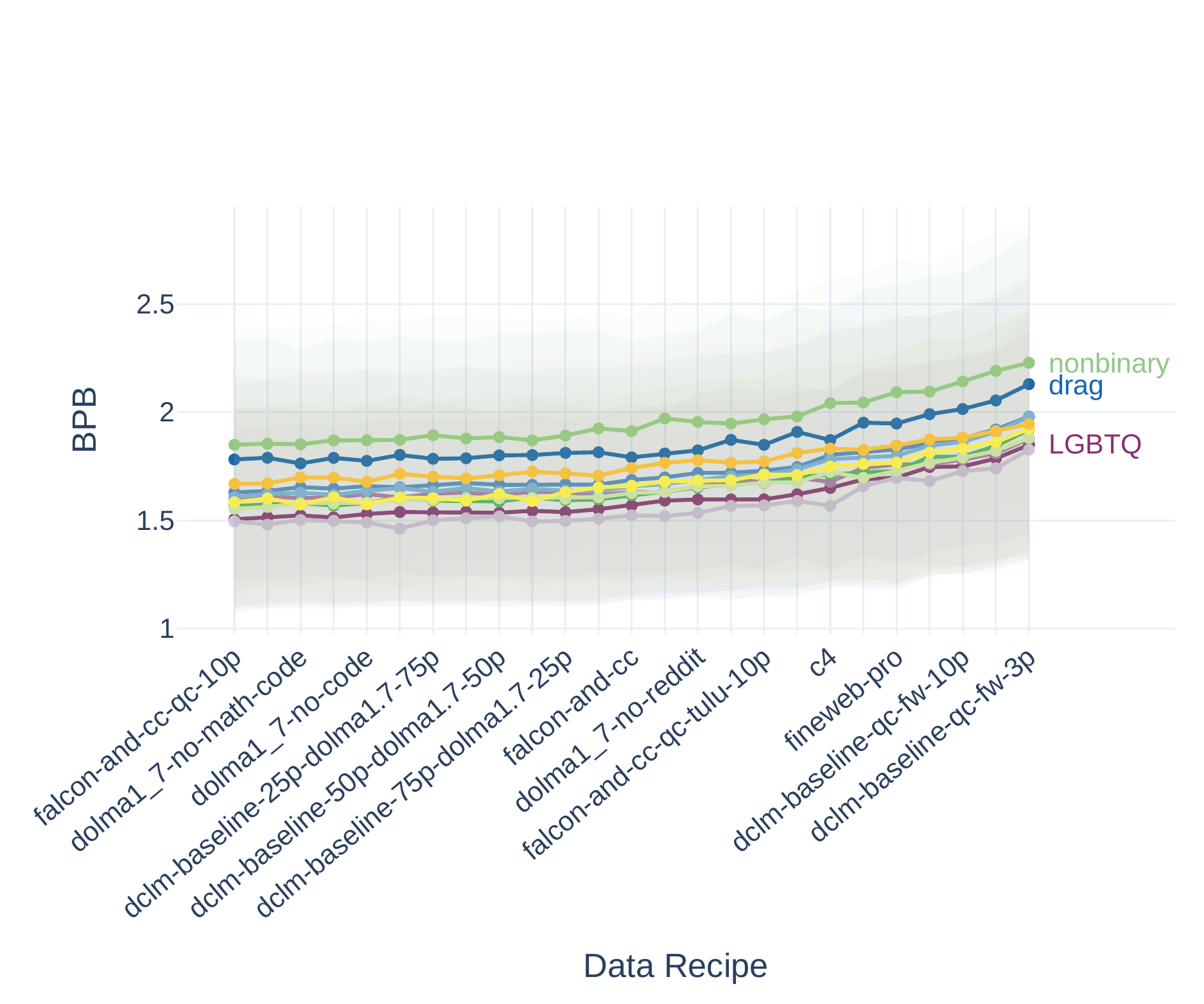}
    \caption{\textbf{Queer subcommunities are not served equally} -- all models perform worse in text containing slang pertaining to non-binary and drag communities. Data recipes contaning a \textsc{CommonCrawl} mix perform generally better across all communities when compared to recipes containing a \textsc{FineWeb} mix.}
    \label{fig:datadecide_bpb_recipes_community}
  \end{subfigure}
  \hfill
  \begin{subfigure}{0.48\textwidth}
    \centering
    \includegraphics[width=\textwidth]{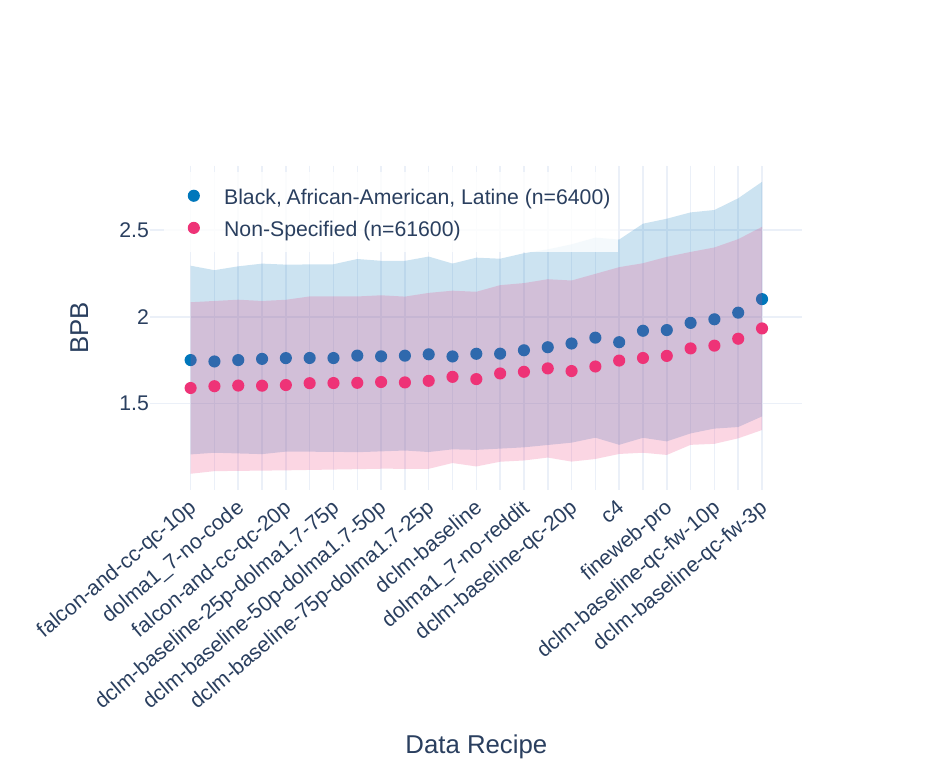}
    \caption{Results for examples containing slang that pertains to African-American or Latine communities, as opposed to non-specified slang. Across all data recipes, \textbf{performance is worse for non-white queer slang.}}
    \label{fig:datadecide_bpb_intersectional}
  \end{subfigure}
  \caption{Average \textsc{Slaying} BPB  for different queer subcommunities, across \textsc{DataDecide} recipes.}
  \label{fig:combined}
\end{figure}

As the primary sources of vocabulary for \textsc{Slaying}, GSSO and Wiktionary label slang terms as pertaining to specific queer subcommunities. As such, \textsc{Slaying} allows us to perform a fine-grained analysis of the slang of transgender, gay, lesbian, and other communities, rather than treating queer vernacular as a monolith. Importantly, a subset of slang terms pertain to specific ethnic communities (African-American, Black, and Latine), allowing for an intersectional analysis of race and queerness. 

\subsubsection{Language Model Fit}

For these purposes, we continue the perplexity-based analysis of \Cref{subsec: Case Study I}. We select queer subcommunities containing at least 5 slang terms, and 10 instances in \textsc{Slaying}, and calculate BPB across \textsc{DataDecide} data recipes (\Cref{fig:datadecide_bpb_recipes_community}). Across all models, performance is worse for slang pertaining to nonbinary and drag communities -- \emph{queer subcommunities are not served equally}. Focusing on the nonbinary community, we can establish a parallel between poor performance on nonbinary slang and other forms of linguistic bias in language models against nonbinary individuals, namely misgendering \citep{dev2021harms,hossain2023misgendered} and/or the inability to correctly process gender-neutral pronouns \citep{gautam2024robust}. Focusing on the drag community, we highlight that many terms from the \texttt{drag slang} class of GSSO (and consequently \textsc{Slaying}), such as \textit{realness}, \textit{shade}, or \textit{slay}, were coined by African-American drag and trans communities. This suggests further linguistic bias against the vernacular of non-white queer communities, which we confirm by measuring BPB on \textsc{Slaying} examples containing slang terms pertaining to Black, African-American, and Latine communities (\Cref{fig:datadecide_bpb_intersectional}). Across all data recipes, we find that performance is worse for non-white slang, relative to terms not specified for ethnicity.

\subsubsection{Toxicity Classification}

\begin{figure}[]
    \centering
    \includegraphics[scale=0.4]{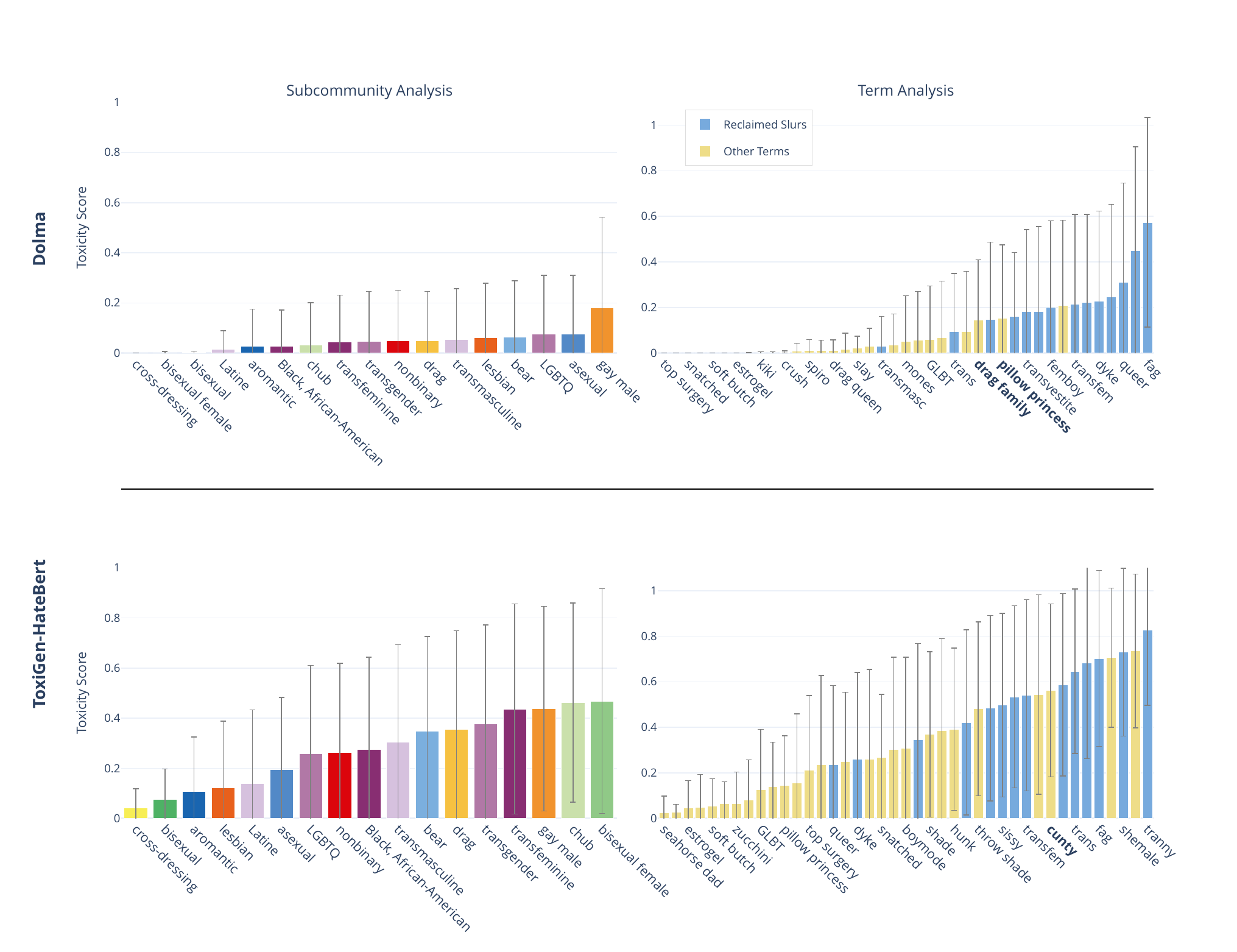}
    \caption{Toxicity of \textsc{Slaying} examples per subcommunity and per term, scored by the \textsc{Dolma} toxicity classifier and \textsc{ToxiGen}'s HateBERT. While results vary across models, a common aspect is that terms other than reclaimed slurs achieve a high toxicity score. \textbf{Different features of queer slang are at risk of being overtly and/or wrongly labeled as toxic}. Many of these terms (\textit{drag family}, \textit{throw shade}) are drag slang, historically associated with African-American and Latine communities.}
    \label{fig:toxicity_communities}
\end{figure}

Another vehicle to understand how different queer subcommunities are served in the current NLP landscape is toxicity classification -- specifically, wrong or overt classification of queer speech as toxic by hate-speech classifiers, a common concern of queer NLP research \citep{dorn2024harmful, weber2026queer}. We leverage the \textsc{Dolma} FastText hate-speech classifier (part of its content filtering pipeline), trained on Jigsaw Toxic Comments \citep{soldaini2024dolma, jigsaw-toxic-comment-classification-challenge}; and \textsc{ToxiGen}'s HateBERT due to its high performance on implicit hate-speech detection tasks \citep{hartvigsen2022toxigen}.

The \textsc{Dolma} classifier and HateBERT, respectively, classify $7.81\%$ and $31.16\%$ of \textsc{Slaying} as toxic, i.e., above the score threshold of $0.5$ -- this is effectively a false positive rate, since all examples of \textsc{Slaying} were classified as harmless. As these results are not particularly informative, we calculate subcommunity-wise and term-wise results, depicted on \Cref{fig:toxicity_communities}. Results vary greatly across models -- the \textsc{Dolma} classifier classifies \textit{gay male} slang as the most toxic, while for HateBERT it is \textit{bisexual female} slang. However, there is an important common aspect: while examples containing reclaimed slurs are labeled as the most toxic, instances of other terms can be equally labeled as toxic by these classifiers. These terms represent different features of queer language and slang formation. Many are drag slang (\textit{throw shade}, \textit{drag family}), historically associated with African-American and Latine communities -- again we see heightened bias against these. Other terms were coined through semantic extension (\textit{cunty}, formed also through adjectivization of \textit{cunt}). Here, we hypothesize that the models are interpreting the term in the conventional sense (the mainstream misogynistic slur \textit{cunt}) rather than the queer slang, reclaimed sense (\textit{amazing or very good or cool, especially with regard to femininity}). Further results regarding this case study, including a more direct comparison with the work of \citet{dorn2024harmful}, can be found in the Appendix \Cref{app:Details of Case Study II}.

\section{Conclusion}

In this manuscript, we introduce \textsc{Slaying}, the first manually annotated dataset reflecting real-world usage of queer slang. We describe its extensive filtering and annotation pipeline, which includes community-driven crowdsourced annotation work -- effectively validating the correctness and safety of the released data. We showcase possible use cases for \textsc{Slaying}, of interest to the NLP fairness research community, and the broader ML field: (i) as a resource to measure linguistic bias -- where we find that models can be unbiased against queer communities while still performing poorly on its slang; and (ii) as a vehicle for analysis of bias against specific queer subcommunities and terminology, finding that models perform worse on slang pertaining to African-American and Latine communities -- terms historically associated with these communities are also more at risk of being overtly labeled as toxic by toxicity classifiers. We hope that the present findings contribute towards fairer NLP systems; and that \textsc{Slaying} can serve as a valuable resource of the ever-changing and rich language of queer people.

\section{Limitations \& Ethical Considerations}
\label{sec:Limitations}

\paragraph{Misuse} The authors do not condone usage of the \textsc{Slaying} dataset to enforce or reinforce stereotypes against the LGBTQ+ community. Specifically, terminology or sentences present in \textsc{Slaying} should not be used to promote transphobic, homophobic, racist, misogynistic, or otherwise bigoted sentiment and talking points. 

\paragraph{Intersectionality and Linguistic Appropriation} Over the course of this work, we refer to \textsc{Slaying} as a dataset of \textit{queer slang}. This decision is not without its caveats. First, we must acknowledge that many of the terms we have collected have their origins in African American English (AAE), often with roots in African-American gay/trans communities and ballroom, and there is a wider conversation around appropriation of these terms and erasure of the linguistic history of these communities \citep{crowley2025and}. Secondly, we recognize that many of these terms are derived from very specific sociolects, without much traction outside of specific queer subcommunities (see, for example, transgender slang originating on 4chan \citep{kotajarvi1998group}).

\paragraph{English Only} This work only considered English queer slang terminology and data. Queer language and experiences vary across languages and cultures. Furthermore, queer slang in predominantly English-speaking countries, such as the USA, can be and is influenced by other languages \citep{morgan2017lesbian}. Multilingual explorations of the intersection of queer language and NLP are a possible avenue for future work. 

\paragraph{Subjectivity \& Generalization of Findings} Labeling for harmful content is inherently subjective due to the nature of the task. Furthermore, the majority of \textsc{Slaying} sentences are sourced from Reddit, therefore potentially being subject to the same demographic skews as Reddit itself \citep{weld2024making}. To mitigate issues regarding subjectivity and selection bias, we ensure total coverage of terms by crowdsourced annotators, and make sure \textsc{Slaying} is open to future revisions. Finally, the present experiments are limited to suites of small models and classifiers. It is unclear how findings extrapolate to larger language models and AI systems.

\begin{ack}
This research was supported by the Munich Center for Machine Learning (MCML) and German Research Foundation (DFG, grant SCHU2246/14-1). We would like to thank Dawar Hakimi, Sebastian Gerstner, Yihong Liu, Ali Modarressi, Clair Kronk, Sophie Hao, Joshua Tint, Jacob Hobbs, and the Queer in AI community for valuable discussions and feedback.
\end{ack}

\bibliographystyle{plainnat} % or use IEEEtran, 
\bibliography{custom}

%%%%%%%%%%%%%%%%%%%%%%%%%%%%%%%%%%%%%%%%%%%%%%%%%%%%%%%%%%%%

\appendix

\section{Detailed Related Work \& Theoretical Background}
\label{app:Related Work and Background: Details}

\paragraph{Queer Linguistics} The latter half of the 20th century saw a rise in research
on gay and lesbian languages. Much of this research has
historically consisted of glossaries of ingroup terms and
discussions of possible etymologies of queer words, and it
has been criticized for overly relying on sexual identity
labels as a foundation for sociolinguistic analysis
\cite{kulick2000gay}. The rise of queer linguistics in the
1990s and 2000s was accompanied by a growing interest in
gender and sexual non-normativity, i.e., perspectives on
queer language as the discursive materialization of all
sexual identities, a tool for identity construction outside
of cisheteronormativity
\citep{motschenbacher2011taking}. However, these
foundational works in queer linguistics have also received
criticism from a transgender linguistics perspective, for
misconstruction of trans identities and non-normatively
gendered communities \citep{zimman2018transgender}. Informed
by collective experiences, specific groups within the queer
umbrella possess different languages. Notably,
\citet{zimman2018transgender} highlights trans discourse as
being especially concerned with authenticity and with institutional powers as tools for oppressing trans individuals. 

\paragraph{Slang in NLP} Inclusion of slang terms in the NLP pipeline has been shown to be beneficial for tasks such as sentiment analysis \citep{wu2018slangsd} and sarcasm detection \citep{wilson2020urban}. Knowledge of slang can also aid LLMs in navigating user interaction. For this purpose, \citet{mei2024slang} design a benchmark for measuring LLMs’ comprehension of newly-emerging idiomatic expressions and slang. \citet{sun2024toward} provide the first comprehensive probing analysis of LLMs on slang knowledge, establishing a benchmark for informal language processing.

\paragraph{NLP for Queer Stakeholders} As noted in \Cref{sec:Background and Related Work}, the subfield of NLP that centers queer individuals and communities as stakeholders of NLP systems is sometimes referred to as ``Queer NLP''. Notably, \citet{weber2026queer} survey these works, highlighting current trends and gaps. The concerns of queer NLP are broad, including misgendering by NLP and AI systems \citep{hossain2023misgendered, gautam2024robust}, and perpetuation of stereotypes \citep{felkner2023winoqueer}. Many of these works propose new resources to mitigate allocational and representational harms, allowing for evaluation of NLP systems. Highlighting some examples that showcase the diversity of queer NLP: \citet{vasquez2022heterocorpus} create a corpus for detection of heteronormative language; \citet{lu2022subtle} create a dataset to detect trans-exclusionary language, while training classifiers to identity this form of toxic speech; and \citet{dacon2022detecting} a introduce a real-world dataset on harmful online speech against queer individuals.

\paragraph{Sociolects \& Linguistic Bias} A rising concern in NLP research is the ability of language models to process linguistic varieties, including sociolects -- of which, notably, African American Vernacular English (AAVE) \citep{groenwold2020investigating, nias2025culture, hofmann2024ai, lin2025assessing}. As noted in \Cref{sec:Limitations}, there is a strong link between queer vernacular and AAVE. The two have been studied in tandem \citep{basoah2025not}.

\paragraph{Digital Archival for Queer Language} Resources such as the Homosaurus \citep{cifor2023mediating}, the Gender, Sex, and Sexual Orientation ontology \citep{kronk2020development}, and lgbtDB\footnote{\url{https://lgbtdb.wikibase.cloud/wiki/Main_Page}} work as a reliable source for queer language (and other queer-related resources). Regardless of these archival efforts, queer-related terms often do not reach crucial corpora, notably those curated to aid hate-speech detection and content moderation. \citet{ramesh2022revisiting} find that English lexicons aimed at content moderation do not make any distinction between pejorative and non-pejorative queer-related words.

\paragraph{Availability of Queer Vocabularies and Resources.} Projects such as \textsc{Slaying} rely on digital archival for queer vocabularies. However, while online content is disappearing at an alarming rate \citep{chapekis2024online}, recent years have seen a rise in disinformation, misinformation, and politically-motivated attacks against individuals, communities, and the availability of queer resources \citep{ellis2023information, hammer2024book, mehra2014don}. While works towards inclusive NLP resources and fair models contribute towards a richer and less cisheteronormative research field, the impact of archival in the preservation of queer history cannot be overstated \citep{rawson2010archiving}. Due to the nature of our vocabulary sources as user-contributed or continuously updated projects, as well as the ever-evolving nature of queer slang, we acknowledge that the term list using in \textsc{Slaying} is not extensive and serves as a snapshot of the currently available resources at the time of writing.

% queer linguistics

 %todo heterocorpus

\section{Sources: Details}

\subsection{Queer Slang Terms}

\subsubsection{Manual Edits \& Deviations from Sources}
\label{app:Manual Edits Deviations from Sources}

In rare cases, terms are marked in our sources as slang terms, but do not include definitions (even after cross-checking with all sources). Here, we list and elaborate on those cases:

\begin{itemize}
    \item \textbf{high femme}: Term occurs in GSSO (as \textit{lesbian slang}). In \textsc{Slaying}, we added the definition \textit{``A femme person who expresses their gender in ultra feminine ways. Often, but not exclusively, this refers to a queer woman.''}, found in the Glossary of The Gender and Sexuality Campus Center\footnote{\url{https://gscc.msu.edu/education/glossary.html}} (Michigan State University).
    \item \textbf{gender straight}: Term occurs in GSSO (as \textit{transgender slang}). We added the definition \textit{``A person who by nature or by choice conforms to gender based expectations of society.''}, found in the glossary of the LGBTQ+ Terminology\footnote{\url{https://www.jjay.cuny.edu/student-life/lgbtq-resource-center/lgbtq-terminology}} resource (John Jay College of Criminal Justice).
    \item{\textbf{gym queen}:} Term occurs in GSSO (as gay male slang). We adapt a definition derived from Wikipedia\footnote{\url{https://en.wikipedia.org/wiki/Queen_(slang)\#Gym_queen}} and the original source of the Wikipedia definition \citep{manso2003ptown}. In \textsc{Slaying}, we define ``gym queen'' as \textit{``Refers to gay men that are into bodybuilding, many of whom are into steroids''}.
    \item{\textbf{femme dyke}: Term occurs in GSSO (as lesbian slang). Unfortunately, a specific and rigorous definition is hard to find at time of writing, and the terms seems nebulous. The term is mostly used a self-identification label\footnote{\url{https://navelgazingwriter.wordpress.com/2018/10/29/en-femme-being-a-femme-dyke/}, \url{https://www.femmedykezine.com/}}. We decided to leave its definition blank and, therefore, the term is not part of the final dataset.} 
    \item{\textbf{transgender until graduation}: Term occurs in GSSO (as \textit{transgender slang}). The term appears to have a small ammount of traction in Internet forums (Reddit, 4chan) but as we were unable to find a rigourous definition, we leave its definition blank.}
    \item{\textbf{mattress muncher}: Term occurs in GSSO (as LGBTQ slang). We find conflicting definitions for ``passive man''\footnote{\url{https://greensdictofslang.com/entry/kqxfp2i}} or ``woman who performs oral sex on another woman''\footnote{\url{https://www.onelook.com/?lang=all&w=mattress+muncher}}.}
\end{itemize}

% \subsubsection{Vocabulary Sources}

% \paragraph{GSSO Details} Due to the very extensive nature of GSSO, we only scrape specifically queer-related slang categories. For this purpose, we collect all classes containing the word ``slang'' in their title, excluding the following due to redundancy ot not being explicitily related to sexual/romantic/gender orientations: slang, polyamory slang, slang dictionary, sexual slang, drug slang, hotel slang, pornography slang, barebacking slang, internet slang, cannabis slang, BDSM slang, sex work slang, and gendered slang.

\subsubsection{Definition Cleaning}
\label{app:Definition Cleaning: Details}

Very common terms (see \textit{bear}, for example) are present and defined by multiple of our sources. This entails that, after scraping, we may have redundant definitions. Nonetheless, there are terms with multiple, distinct, valid definitions. A notable example is \textit{angel}, defined in lgbtDB as both ``\textit{A gay man}'' and as ``\textit{A vogue performer who is perceived as feminine or femme and who is perceived as having a daintier dance style or execution.}'', in the context of ballroom. Our term definitions should be as descriptive and concise as possible, while retaining the valid, distinct ones.

We embed all definitions using the \texttt{all-mpnet-base-v2} sentence embedder, chosen due to its top performance as a general-purpose model\footnote{\url{https://www.sbert.net/docs/sentence_transformer/pretrained_models.html}}. We iterate through all possible definitions for a term, calculating the respective cosine similarity matrix. If two definitions have a similarity score higher than a thresholdn $t$, we discard the shorter one (hypothesizing that the longer description will be more detailed and specific).

We test values for $t$ $min = \{0.60, 0,65, 0.70, 0,75\}$. We settled on $0.7$ as the optimal value (\Cref{tab:similarity_experiments}), since it offers a balance between effective filtering and retaining distinct senses.

Having this in mind, we design and clean our slang term list according to the following:

\begin{itemize}
    \item One entry of our slang term list contains a \textit{term}, optional one or more \textit{variants}, and one or more \textit{definitions}. \textit{Variants} are a catch-all field for synonyms or terms with the same stem. We do not manually adjust our sources in that regard: for Chew, we add other terms with the same definition as variants; for GSSO, we extract the \texttt{SYNONYM} property terms as variants (which can include terms with the same stem, or terms with the same definition). Additionally, terms with defintions consisting only of ``\textit{Synonym of X}'' were re-categorized as variants of X. Note that, duritng the filtering process for the creation of \textsc{Slaying}, there is no difference between \textit{terms} and \textit{variants}. We organize them as such in our slang term list for preservation of the relation between the terms.

    \item We remove extra notes related to archival (such as ``\textit{Widely used, but should be enclosed in quotation marks and contextualized if needed in archival description.}'', from the Chew Glossary) and references to other terms ``\textit{see also: ...}'' while preserving the remainder of the definition. 

    \item Definitions can be highly redundant (``\textit{To behave, speak, act, etc. in a flamboyant, affected, ostentatious, or exaggerated manner stereotypically associated with gay men.}'' and ``\textit{A gay man who is considered to be ostentatious, over the top, theatrical, or flamboyant.}'', for example). We create a sentence embedding of each definition and remove definitions that have a cosine similarity $> 0.7$, keeping the longest of the pair.

    \item We remove completely circular definitions, or definitions that solely define that term by mentioning another term belonging to the slang list (for example, the definition ``\textit{A boydyke.}'' for the term ``lesboy'' is removed).

    \item We adapt definitions that partially define a term by refering to another term pertaning to the slang list. For example, ``\textit{Being or resembling a twunk (“muscular twink”).}'' becomes ``\textit{Being or resembling a twunk (“muscular twink”, "twink" meaning "young, attractive, slim male").}''. We acknowledge that this is only possible to an extent, since very well-known terms (such as ``lesbian'') are widely used in definitions for other terms. 

\end{itemize}

\begin{table}[]
\small
\centering
\caption{Filtered definitions and respective cosine similarity. We choose $0.7$ as a threshold, as it (empirically) offers a good balance between effective filtering and retaining distinct senses.}
\begin{tabular}{@{}p{1cm}p{1cm}p{1cm}p{4.3cm}p{4.3cm}@{}}
\toprule
\textbf{Term} & \textbf{Similarity Value} & \textbf{Similar Senses?} & \textbf{Definition 1 (kept in case of high similarity)} & \textbf{Definition 2 (discarded in case of high similarity)} \\
\midrule
hetty & 0.78 & Yes &
A heterosexual. &
Heterosexual. \\
\midrule
% bean queen & 0.75 & Yes &
% Slang used in male homosexual communities in the USA to describe Mexican homosexuals. Racist overtones. Also used to describe Latinx drag queens, as well as white men who are attracted to Latino men. &
% A non-Latino man who displays fetishistic attraction toward Latino men. \\
% \midrule
% egg & 0.72 & Yes &
% Slang term (originating online) for a person who has not yet realised they are transgender, has not yet come out, or is in the early stages of transitioning. Usually used (fondly) by trans people to recognise when aspects of someone's personality or behavior remind them of gender-related aspects of themselves before they realized they were trans. &
% A transgender person who considers themselves to not be transgender or a transgender person who has not yet made their identity public. When a transgender person comes out, it is said that "the egg cracks". \\
% \midrule
% man pussy & 0.66 & No &
% The anus and rectum of a man, usually the receptive participant (the bottom) in gay sex. &
% The vulva or vagina of a trans man. \\
% \midrule
fairy lady & 0.64 & No & Some sources cite this term to mean a feminine lesbian, others to mean a bisexual person of any gender. Mostly appeared originally in 1920s-onwards in informal dictionaries of slang, which were not always reliable. &
In a lesbian relationship, the partner who takes the passive, receptive, or submissive role during sexual activities.

\end{tabular}
\label{tab:similarity_experiments}
\end{table}

\begin{table}[]
\centering
\caption{Annotation example for the task of sense disambiguation, during the pre-filtering stage of the curation of \textsc{Slaying}. This internal manual annotation effort yields a gold set, used to evaluate LLMs for pre-filtering purposes.}
\small
\begin{tabular}{@{}lp{2cm}p{1.5cm}p{3.5cm}ccccc@{}}
\toprule
\textbf{Term} & \textbf{Definition 1} & \textbf{Sentence} & & \textbf{1} & \textbf{2} & \textbf{3} & \textbf{4} & \textbf{5} \\
\midrule
% Example row (empty):
 example term & example definition & example sentence & Do any of the definitions apply to the term as used in the sentence? & $\square$ & $\square$ & $\square$ & $\square$ & $\square$ \\
\bottomrule
\\
\end{tabular}
\label{tab:prefiltering_annotation}
\end{table}

\subsection{Raw Data}
\label{app: Sources: Details}

\paragraph{Social Media}

The social media subset represents
%57.89\%
58\%
of the total raw dataset (114,596 out of 197,958 examples).
All examples were retrieved from the forum-based social
media platform
Reddit\footnote{\url{https://www.reddit.com/}}, specifically
from subreddits listed in a community-curated
list\footnote{\url{https://www.reddit.com/r/LGBTdirectory/}}. It
includes 792 entries of subreddits related to queer
identities and additional metadata such as thematic
descriptions and community tags. Using the Reddit API, this
list of subreddits was reduced to exclude inactive, private,
and invite-exclusive subreddits, as well as those containing
adult content (18+), yielding 264 subreddits. For each
subreddit, up to 15 of the most relevant posts are retrieved in regard to a specific slang term. From the content of each post, we collect individual sentences from the main body of the post, as well as the first 5 pages of comments. This collection of potential instances is filtered to exclude sentences that include links to webpages,
keywords related to promo codes, sentences consisting of only capital characters or at least 40\% of non-alphanumerical characters, (e)mail addresses, phone numbers, and references to other subreddits or reddit users. Sentences that adhere to these rules, include a slang term, and have between 4 and 30 whitespace-separated tokens are added to the raw dataset. We do not claim ownership of these entries; all property rights remain with Reddit and the original users. Entries are anonymized and this process did not exceed the rate limits of the Reddit API.

\paragraph{Podcasts}

We collected textual data from the publicly available platform Podscripts\footnote{\url{https://podscripts.co}}, restricting the scope to the Society \& Culture category (113 podcasts). We restricted the scope to the Society \& Culture category and retrieved only sentences matching predefined query terms. From these, we retained short excerpts (4 to 30 whitespace-delimited tokens) to ensure concise and contextually meaningful examples. Each occurrence of a target term was normalized to lowercase and stored along with its podcast source. This produced a balanced set of relevant term occurrences from a large, heterogeneous podcast corpus. The podcasts subset represents 35\% of the total raw dataset (70,185 examples). We do not claim ownership of the underlying content; all intellectual property rights remain with the original podcast creators and rights holders. The dataset consists exclusively of minimal, non-substitutive quotations used for non-commercial academic research and does not enable reconstruction of full transcripts or episodes. Data access was limited to publicly available content, and we adhered to responsible data collection practices by minimizing retained text and preserving source attribution.

\paragraph{Subtitles}
From the OpenSubtitles Corpus
\citep{lison2016opensubtitles2016}, we collected subtitles
from prominent queer-related movies and television
shows. The titles were retrieved from IMDb lists
self-described as pertaining to gay, lesbian, transgender,
drag, LGBTQ+, or campy
titles\footnote{\url{https://www.imdb.com/list/ls033185849/},
\url{https://www.imdb.com/list/ls009573275/},
\url{https://www.imdb.com/list/ls055851154/},
\url{https://www.imdb.com/list/ls023525140/},
\url{https://www.imdb.com/list/ls049480201/},
\url{https://www.imdb.com/list/ls541265294/},
\url{https://www.imdb.com/list/ls090541670/}}. From those
titles, we select only those publicly available in the
OpenSubtitles Corpus\footnote{\url{http://www.opensubtitles.org/}} (licensed through ODC-BY). The subtitles subset represents 7\% of the total raw dataset (13,177 examples). Valid sentences consist of between 4 and 30 whitespace-separated tokens.

\section{Prefiltering: Details}
\label{app:Prefiltering: Details}

\subsection{Prompting Setup}

The gold standard set for sense disambiguation (\Cref{tab:prefiltering_annotation}) that is described on \Cref{subsec:Pre-filtering} consists of 6591 examples. From that set, 250 were used to evaluate the previously described model suite, and other 250 examples were to test different prompt variations, on GPT-5 Mini. The latter results are depicted on \Cref{tab:prompt-templates}. We settled on the ``contextual'' prompt (which provides some added context to the task) as the highest performing variation \Cref{tab:prompt-results}. 

We argue that state-of-the-art reasoning models can serve as annotation tools for queer language in a pre-filtering stage. Low agreement with human annotators on a considerable percentage of queer slang terms (as described in \Cref{subsec:Pre-filtering}), as well as ethical concerns regarding using LLMs as a sole annotator for potentially offensive terms, such as reclaimed slurs, make expert and community-driven annotation necessary.

\begin{table}[]
\centering
\caption{Prompt variations for the task of sense disambiguation, as used to test LLMs as pre-filtering models.}
\label{tab:prompt-templates}
\small
\begin{tabular}{@{}lp{10cm}@{}}
\toprule
\textbf{Variation} & \textbf{Prompt Template} \\
\midrule

\texttt{vanilla} & 
Term: "\{term\}" | Definition: "\{definition\}" | Sentence: "\{sentence\}" | Does the given definition apply to the term as used in the sentence? Answer only with the number in the scale, nothing else. | - [ ] 1 (Not at all) - [ ] 2 (Mostly not) - [ ] 3 (Somewhat) - [ ] 4 (Mostly) - [ ] 5 (Completely) \\
\midrule

\texttt{vanilla\_task} & 
Term: "\{term\}" | Definition: "\{definition\}" | Sentence: "\{sentence\}" | Task: Does the given definition apply to the term as used in the sentence? Answer only with the number in the scale, nothing else. | - [ ] 1 (Not at all) - [ ] 2 (Mostly not) - [ ] 3 (Somewhat) - [ ] 4 (Mostly) - [ ] 5 (Completely) \\
\midrule

\texttt{contextual} & 
Context: Queer slang terms can often be ambiguous, since many have common, non-queer related senses (see "mother", or "read"). In the following task you will be given a queer slang term, one or more respective definitions, and a sentence where that term is used. Your task is to decide whether the meaning of the term in the sentence matches the given definition. | Term: "\{term\}" | Definition: "\{definition\}" | Sentence: "\{sentence\}" | Task: Does the given definition apply to the term as used in the sentence? Answer only with the number in the scale, nothing else. | - [ ] 1 (Not at all) - [ ] 2 (Mostly not) - [ ] 3 (Somewhat) - [ ] 4 (Mostly) - [ ] 5 (Completely) \\
\midrule

\texttt{one-shot} & 
Example: Term: "throw shade" | Definition: "To subtly insult someone." | Sentence: "She loves to throw shade at her ex." | Task: Does the given definition apply to the term as used in the sentence? Answer only with the number in the scale, nothing else. | - [ ] 1 (Not at all) - [ ] 2 (Mostly not) - [ ] 3 (Somewhat) - [ ] 4 (Mostly) - [ ] 5 (Completely) | 5 | Now evaluate the following: | Term: "\{term\}" | Definition: "\{definition\}" | Sentence: "\{sentence\}" | Task: Does the given definition apply to the term as used in the sentence? Answer only with the number in the scale, nothing else. | - [ ] 1 (Not at all) - [ ] 2 (Mostly not) - [ ] 3 (Somewhat) - [ ] 4 (Mostly) - [ ] 5 (Completely) \\
\midrule

\texttt{contrastive} & 
The same term can mean different things in queer slang vs. standard English. | Term: "\{term\}" | Definition: "\{definition\}" | Sentence: "\{sentence\}" | Task: Does the given definition apply to the term as used in the sentence? Answer only with the number in the scale, nothing else. | - [ ] 1 (Not at all) - [ ] 2 (Mostly not) - [ ] 3 (Somewhat) - [ ] 4 (Mostly) - [ ] 5 (Completely) \\

\bottomrule
\end{tabular}
\end{table}

\begin{table}[]
\centering
\caption{Analysis of different prompt variations for the task of sense disambiguation, tested with GPT-5 Mini on a sample of 250 raw data examples. The \textit{contextual} prompt shows a higher F1 score (superior alignment with human annotators) and a higher ratio of terms with a high performance. If $F1 > 0.7$ or $\alpha > 0.6$ for the instances of a term, we consider that the model has ``high performance'' on that term.}
\begin{tabular}{lccc}
\toprule
Prompt & High Performance Ratio & F1 Score & Krippendorff's $\alpha$ \\
\midrule
contextual & \textbf{0.855} & \textbf{0.864} & \textbf{0.152} \\
contrastive & 0.834 & 0.842 & 0.110 \\
one-shot & 0.829 & 0.841 & 0.137 \\
vanilla task & 0.824 & 0.832 & 0.112 \\
vanilla & 0.829 & 0.845 & 0.122 \\
\bottomrule
\\
\end{tabular}
\label{tab:prompt-results}
\end{table}

% \subsection{Validating GPT-5 Mini as a Pre-filtering Model}
% \label{app:Validating GPT-5 Mini as a Pre-filtering Model}

% To assure that these annotators have a high agreement, they annotated an extra shared subset of 25 examples, yielding a Krippendorff's $\alpha$ of $0.89$. 

% We hypothesize that the performance of LLMs as classifiers for the task of sense disambiguation can vary greatly per term. As such, to validate GPT-5 Mini as a pre-filtering model, we create a data subset consisting of up to 15 examples per term -- yielding 6591 examples. These were split across three internal annotators, each annotating $ \approx$ 2190 examples. GPT-5 Mini achieves micro F1 scores of 0.76, 0.79, and 0.69, for the three annotators. We argue that this validates the model as a pre-filterer for sense disambiguation.

\subsection{Data to Manually Annotate}
\label{app:Data to Manually Annotate}

%9\% of this sample was sourced from OpenSubtitles, 10\% from Podcasts, and 81\% from Reddit, covering 506 terms.

After pre-filtering the raw version of \textsc{Slaying} with GPT-5 Mini, we establish desiderata for the cleaned data we wish to manually annotate:

\begin{itemize}
    \item We include examples that were classified as correct for the sense disambiguation task (5, meaning the given definition applies \textit{completely} to the term as used in the sentence, as depicted in \Cref{tab:prefiltering_annotation} and \Cref{tab:prompt-templates}). If there are no instances of a term being classified as a 5 we include its instances that were classified as 4 and 3 -- this is a minimal further annotation effort that can result in higher coverage of terms.
    \item We deduplicate very similar examples -- using the \texttt{all-mpnet-base-v2} embedder, we remove examples with a similarity $>0.85$ to any other example, keeping the longest of the pair. This step is ignored if the term has 10 or less entries, to assure vocabulary coverage.
    \item Examples were balanced regarding sources, except for potentially offensive terms -- we assume these are more likely to be used in a reclaimed fashion on Reddit \citep{cepollaro2022reclaims, dorn2024harmful}, and we want to prioritize coverage of these terms.
\end{itemize}

\section{Annotation: Details}
\label{app:Annotation: Details}

Each PA batch contained 1000, 2000, and 2748 examples -- the increased length was validated by a stable high agreement, depicted in \Cref{fig:internal_stability}. 

\begin{figure}
    \centering
    \includegraphics[scale=0.25]{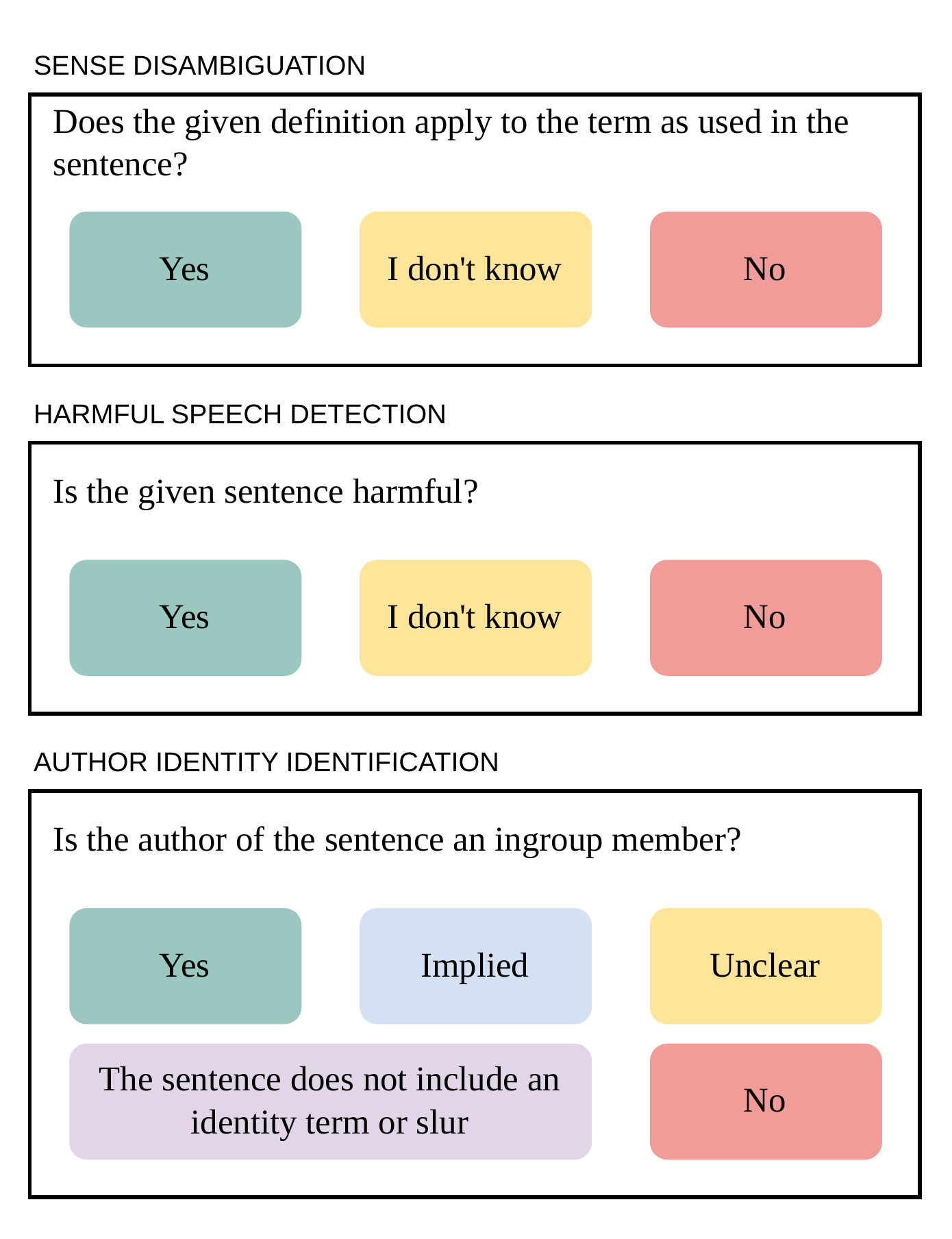}
    \caption{Classification labels for each \textsc{Slaying} annotation task, used for every step of the annotation pipeline.}
    \label{fig:annotation_tasks}
\end{figure}

\subsection{Inter-annotator Agreement Metrics}
\label{app:Inter-annotator Agreement Metrics}

\paragraph{Percent Agreement} Represents the proportion of items for which annotators assign the same label. As it does not account for agreement that could occur by chance, 
$P_a$ can overestimate reliability in cases of data imbalance. It is still widely reported, usually along chance-corrected metrics \citep{abercrombie2023temporal, james2026counting}.
$$P_a = \frac{\text{Number of agreements}}{\text{Total number of observations}}$$

\paragraph{Cohen's $\kappa$} Is a measure of agreement between two raters who express their judgments on a nominal scale \citep{cohen1960coefficient}. Cohen's $\kappa$ accounts for chance agreement, and is therefore considered more reliable than $P_a$. If there are more than two raters, but each unit has been rated by the same number of raters, Fleiss' $\kappa$ may be appropriate \citep{rau2021evaluation}. Cohen's $\kappa$ is defined as
$$\kappa = \frac{p_o - p_e}{1 - p_e}$$
where $p_o$ is the observed agreement and $p_e$ is the hypothetical probability of chance agreement:
$$p_e = \sum_k{\widehat{p_{k12}}}$$
Where $\widehat{p_{k12}}$ is the estimated probability that both rater 1 and rater 2 will classify the same item as $k$.

\paragraph{Gwet's AC1} In cases of data imbalance, $\kappa$ coefficients occasionally yield unexpected results in situations known as the paradoxes of kappa -- where $\kappa$ can
be low despite high observed agreement, or vice-versa. To mitigate this, \citet{gwet2008computing} proposed AC1, which not depend upon the hypothesis of independence between raters, and has been shown to be more stable than Cohen's $\kappa$ \citep{wongpakaran2013comparison}. It has seen increased usage in NLP works, especially those with difficult or highly subjective annotation tasks \citep{james2026counting, bianchi2022s}. Due to the high level of data imbalance present in \textsc{Slaying}, we use Gwet's AC1 as one of our primary metrics. Is it defined by 
$$AC1 = \frac{p_o - p_e^{AC1}}{1 - p_e^{AC1}}$$
where $p_o$ is the observed agreement and:
$$p_e^{AC1} = \frac{1}{N-1} \sum_{i=1}^{N} \pi_i (1 - \pi_i)$$
with $\pi_i$ being the probability that a rater classifies an item into category $i$, and $N$ being the number of categories.

\paragraph{Fleiss's $\kappa$} Generalizes $\kappa$ beyond a pair of raters \citep{fleiss1971measuring}. It also takes into account chance agreement, but it
aggregates agreement across all annotators rather than comparing them pairwise \citep{james2026counting}. It is defined as
$$\kappa = \frac{\bar{P} - P_e}{1 - P_e}$$
where $\bar{P}$ is the mean observed agreement across all $N$ items, defined as:
$$\bar{P} = \frac{1}{N} \sum_{i=1}^{N} P_i$$
and $P_i$ is the observed agreement for each item $i$ among $A$ annotators, defined as:
$$P_i = \frac{1}{A(A-1)} \sum_{j=1}^{k} n_{ij}(n_{ij} - 1)$$
where $n_{ij}$ is the number of annotators who assigned item $i$ to category $j$. The expected agreement $P_e$ is derived from the overall category proportions.

\paragraph{Krippendorff's $\alpha$} Works for various data types (nominal, ordinal, interval), any number of annotators, and can handle missing data. Similarly to Cohen's $\kappa$, it is also not meaningful for cases where all annotators assigned the same label. Is is defined by
$$\alpha = 1 - \frac{D_o}{D_e}$$
where $D_o$ is the observed disagreement and $D_e$ is the expected disagreement. For nominal data (the case in \textsc{Slaying}), disagreement is usually defined as 0 if two labels are the same \citep{james2026counting}.

\subsection{Annotation Sheet and Guidelines}
\label{app: Annotation Sheet and Guidelines}

%The first author (which serves as the principal annotator, henceforth PA) designed the annotation guidelines. These guidelines were used in a pilot study, where the PA and three other authors (henceforth IAs, or internal annotators) label a sample of 25 examples. This effort corresponded to an average pairwise percent agreement of 0.75 (sense disambiguation), 0.78 (harmful speech detection), and 0.53 (author identity identification). After discussion of disagreements and refinement of the guidelines, the agreement increased to 0.81, 0.88, and 0.75. 

The annotation guidelines fined-tuned during the internal annotation process, and used for crowdsourced annotators, are depicted below.

\begin{tcolorbox}[breakable]

You will be labeling examples of usages of queer slang terms. Each example will contain a
slang term, one or more definitions for that term, and a sentence where that term is being
used. The examples you will label may contain slurs, or potentially offensive language.
Your tasks will be highly subjective, particularly for short sentences that do not offer much context -- this is exactly why we need your help! If you are extremely unsure, simply respond \texttt{I don't know / Unclear} . If there is an observation or note you want to add about that particular example and your decision, you can do so in the \texttt{Notes} field!

Your role is to label the examples for the following 3 tasks:

\textbf{Task \#1: Sense Disambiguation} 

\textbf{Task Definition} \\
Given a definition for a term, identify whether that it applies to the term as used in the given sentence.

A term may have multiple definitions (in your annotation sheet, they will be split by bullet points). If that is the case, the term as used in the sentence needs only to satisfy one of the definitions.

\textbf{EXAMPLE 1}\\
Term: ``slay''\\
Definition: ``Used to express approval or amazement.''\\
Sentence: ``Does the outfit slay or no?''\\

Does the given definition apply to the term as used in the sentence?\\
\XBox Yes\\
$\Box$ No \\
$\Box$ I don't know \\

\textbf{EXAMPLE 2}\\
Term: ``read'' \\
Definition: ``To call attention to someone's flaws in a playful, taunting, or insulting manner; alternatively, to recognize someone as being transgender (to clock)..'' \\
Sentence: ``I read a great book yesterday!''

Does the given definition apply to the term as used in the sentence?\\
$\Box$ Yes\\
\XBox No \\
$\Box$ I don't know \\

\textbf{Task \#2: Harmful Speech Detection} 

\textbf{Task Definition} \\

A sentence should be considered harmful if it:
1. uses a slur to derogate and/or dehumanize (e.g., "I think queers are disgusting"), as
opposed to a non-derogatory usage (which includes reclaimed/appropriative usages,
discussion regarding the slur, or recollection of the time the slur was used against the
author -- e.g., "My classmates always called me a queer, I guess they were right!" ). 

2. attacks, seeks to silence, or negatively stereotypes a minority (e.g., "Gay men have such
annoying voices!").

3. criticizes a minority without a well founded argument.

4. blatantly misrepresents truth or seeks to distort views on a minority with unfounded
claims (e.g., "Queer people make terrible parents").

5. promotes, but does not directly use, hate speech or violent crime.

6. causes another person reasonable fear of personal injury.

7. defends an oppressive system, such as xenophobia, sexism, homophobia, or transphobia
(e.g., "Trans men are just confused women").

\textbf{EXAMPLE 1}\\
Term: ``dyke'' \\
Definition: ``Contested terminology. Slur for lesbian, highly offensive if applied to a lesbian or
transmasc person externally. Term has been reclaimed by some lesbians communities, and is
sometimes used affirmingly within communities."Bulldyke" and "bulldagger" are often used
to indicate butch or masculine queer women, sometimes with overtones of toughness or
aggression.'' \\
Sentence: ``I love hanging out at my local dyke bar!''

Is the given sentence harmful?\\
$\Box$ Yes\\
\XBox No \\
$\Box$ I don't know \\

\textbf{EXAMPLE 2}\\
Term: ``faggot'' \\
Definition: ``A pejorative term used chiefly in North America primarily to refer to a gay man.
Alongside its use to refer to gay men in particular, it may also be used as a pejorative term for
a `repellent man' or to refer to women who are lesbian (or to transgender people, typically
transgender women). Its use has spread from the United States to varying extents elsewhere
in the English-speaking world through mass culture, including film, music, and the Internet.''
Sentence: ``I'm ok with gays, but I hate faggots''

Is the given sentence harmful?\\
\XBox Yes\\
$\Box$ No \\
$\Box$ I don't know \\

\textbf{Task \#3: Author Identity Identification} 

\textbf{Task Definition} \\

In a sentence with an identity term or slur, we say the ingroup is the population referenced
by the identity term or slur’s neutral correlate (e.g. the neutral correlate for `dyke' is
`lesbian'). The outgroup is the population not referenced by the identity term or neutral
correlate.
Do you believe the sentence indicates the author is a member of the ingroup? This is a
highly subjective task, so do your best! In case of exteme doubt or if the sentence is
extremely unclear, simply label it as \texttt{Unclear}. Keep in mind that not all slang terms are not
identity terms or slurs (such as ``slay'', ``tea'', etc...) -- in those cases, simply label it as \texttt{The sentence does not include an identity term or slur}.
If you are unsure if a term is an identity term, ask yourself if you can refer to an individual
as that term -- if so, then it is!

\textbf{EXAMPLE 1}\\
Term: ``femme'' \\
Definition: ``Traditionally, a lesbian who appears `feminine' or acts in a `feminine' manner.
However, the term is often used to describe a wide variety of gender expressions and
identities adjacent to feminine or femme-ness.'' \\
Sentence: ``I identify as a femme.''

Is the author of the sentence an ingroup member?\\
$\Box$ The sentence does not include an identity term or slur\\
\XBox Yes \\
$\Box$ No \\
$\Box$ Implied \\
$\Box$ Unclear \\

\textbf{EXAMPLE 2}\\
Term: ``bear'' \\
Definition: ``Community term with multiple meanings. 1. A gay or bisexual man who has
facial/body hair and a cuddly body. 2. An umbrella term that refers to members of a
subculture in the gay and bisexual male communities and is often defined as more of an
attitude or sense of comfort with natural masculinity and bodies.'' \\
Sentence: ``I went to a bear bar yesterday, and had a lot of fun!''

Is the author of the sentence an ingroup member?\\
$\Box$ The sentence does not include an identity term or slur\\
$\Box$ Yes \\
$\Box$ No \\
\XBox Implied \\
$\Box$ Unclear \\

\textbf{EXAMPLE 3}\\
Term: ``slay'' \\
Definition: ``Used to express approval or amazement.'' \\
Sentence: ``He was really slaying that outfit.'' 

Is the author of the sentence an ingroup member?\\
\XBox The sentence does not include an identity term or slur\\
$\Box$ Yes \\
$\Box$ No \\
$\Box$ Implied \\
$\Box$ Unclear \\

\end{tcolorbox}

\subsection{Public \& Private \textsc{Slaying}}

\begin{figure}[]
    \centering
    % First row - 2 images
    \begin{subfigure}[b]{0.48\textwidth}
        \centering
        \includegraphics[width=\textwidth]{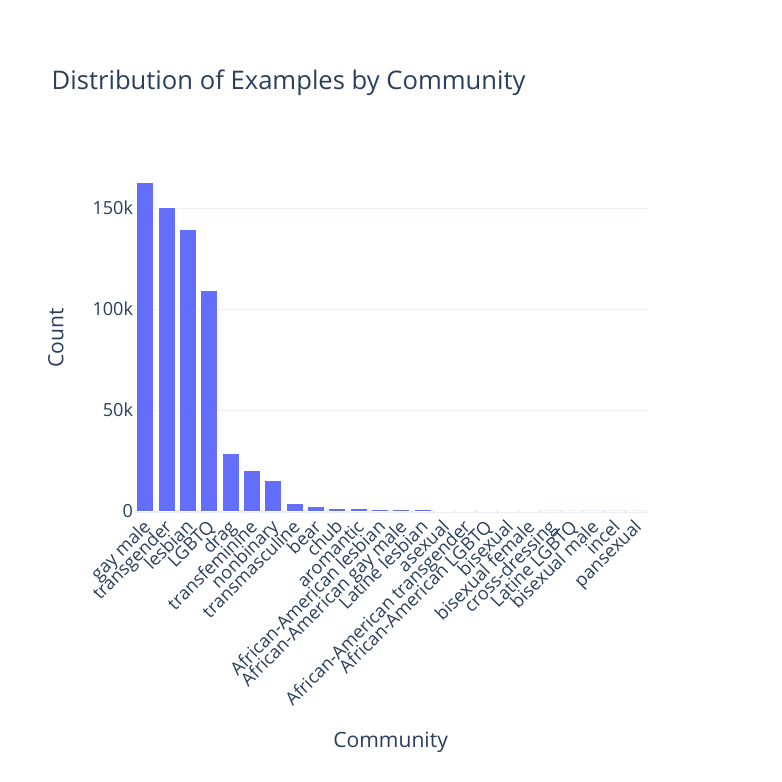}
        \label{fig:sub1}
    \end{subfigure}
    \hfill
    \begin{subfigure}[b]{0.48\textwidth}
        \centering
        \includegraphics[width=\textwidth]{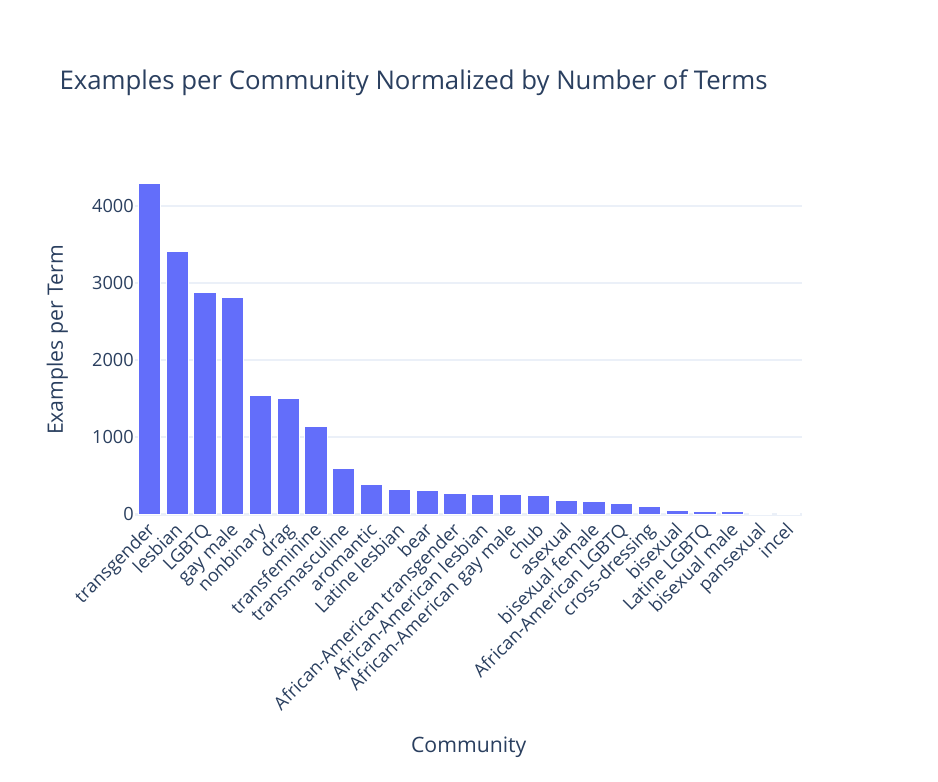}
        \label{fig:sub2}
    \end{subfigure}
    
    \vspace{0.5em}
    
    % Second row - 2 images
    \begin{subfigure}[b]{0.48\textwidth}
        \centering
        \includegraphics[width=\textwidth]{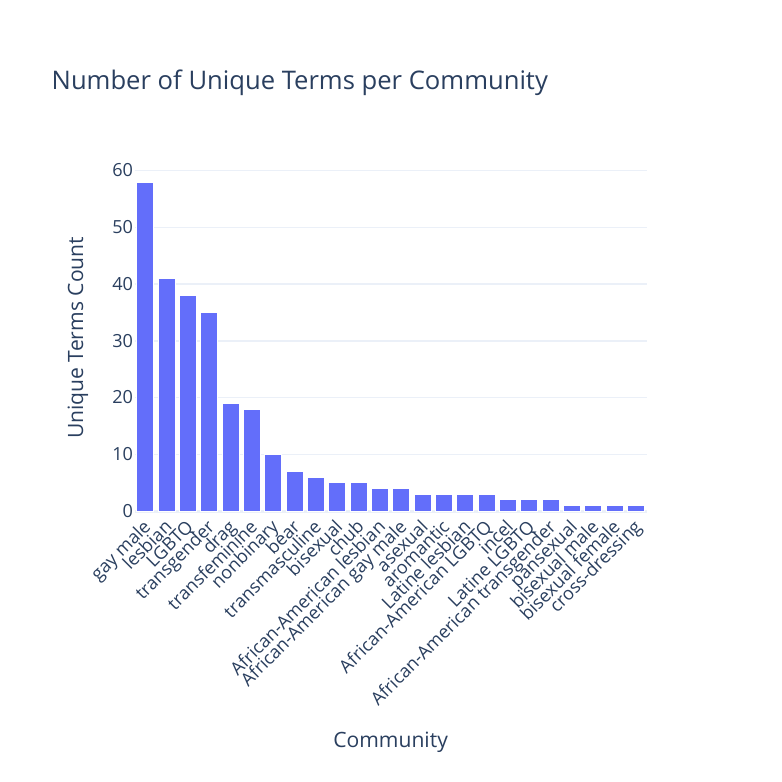}
        \label{fig:sub3}
    \end{subfigure}
    \hfill
    \begin{subfigure}[b]{0.48\textwidth}
        \centering
        \includegraphics[width=\textwidth]{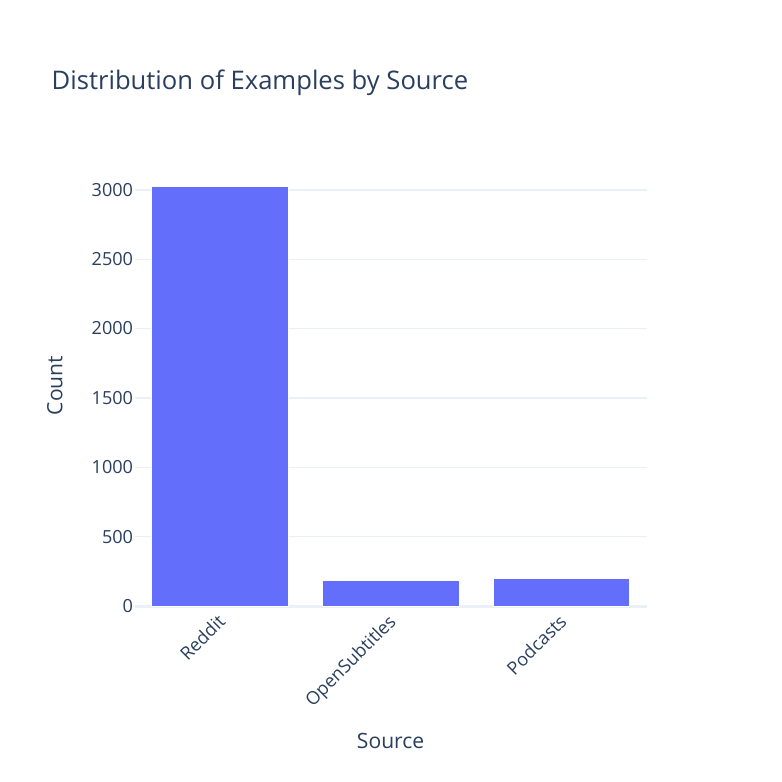}
        \label{fig:sub4}
    \end{subfigure}
    
    \vspace{0.5em}
    
    % Third row - 1 wide image
    \begin{subfigure}[b]{\textwidth}
        \centering
        \includegraphics[width=\textwidth]{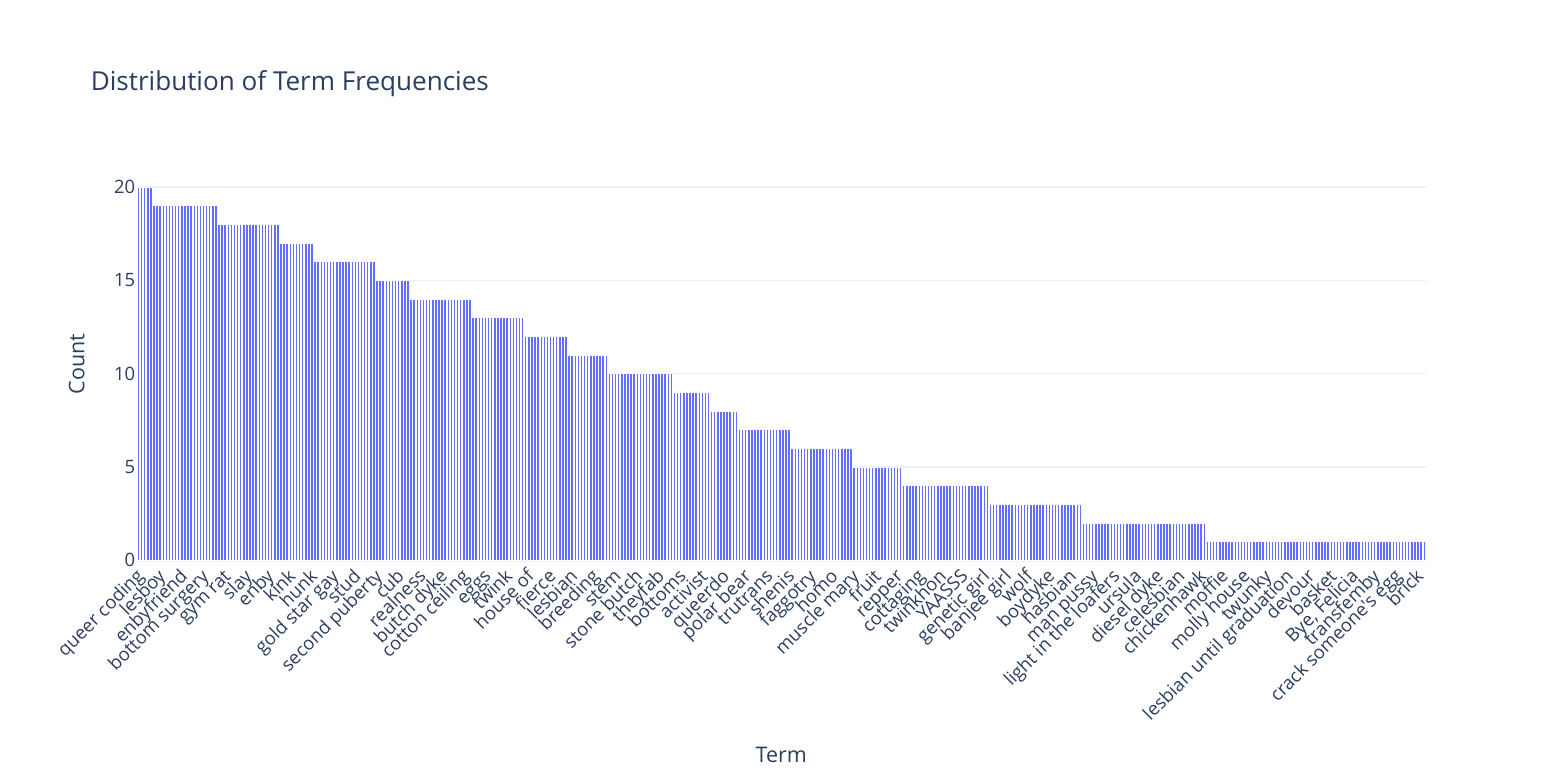}
        \label{fig:sub5}
    \end{subfigure}
    
    \caption{Distribution of terms, communities, and sources for the public version of \textsc{Slaying}.}
    \label{fig:public_stats}
\end{figure}

% priv slaying stats
\begin{figure}[]
    \centering
    % First row - 2 images
    \begin{subfigure}[b]{0.48\textwidth}
        \centering
        \includegraphics[width=\textwidth]{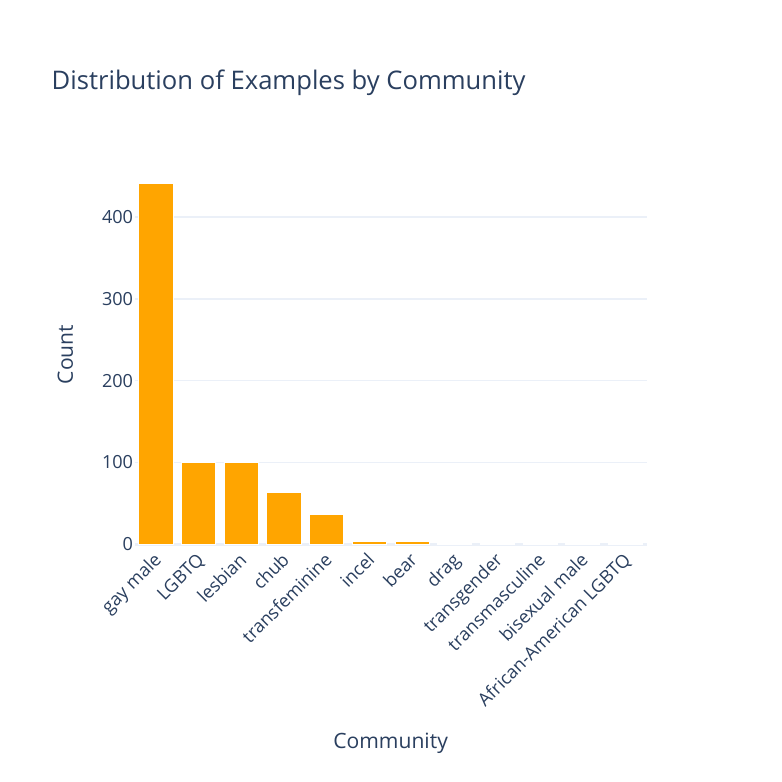}
        \label{fig:sub1}
    \end{subfigure}
    \hfill
    \begin{subfigure}[b]{0.48\textwidth}
        \centering
        \includegraphics[width=\textwidth]{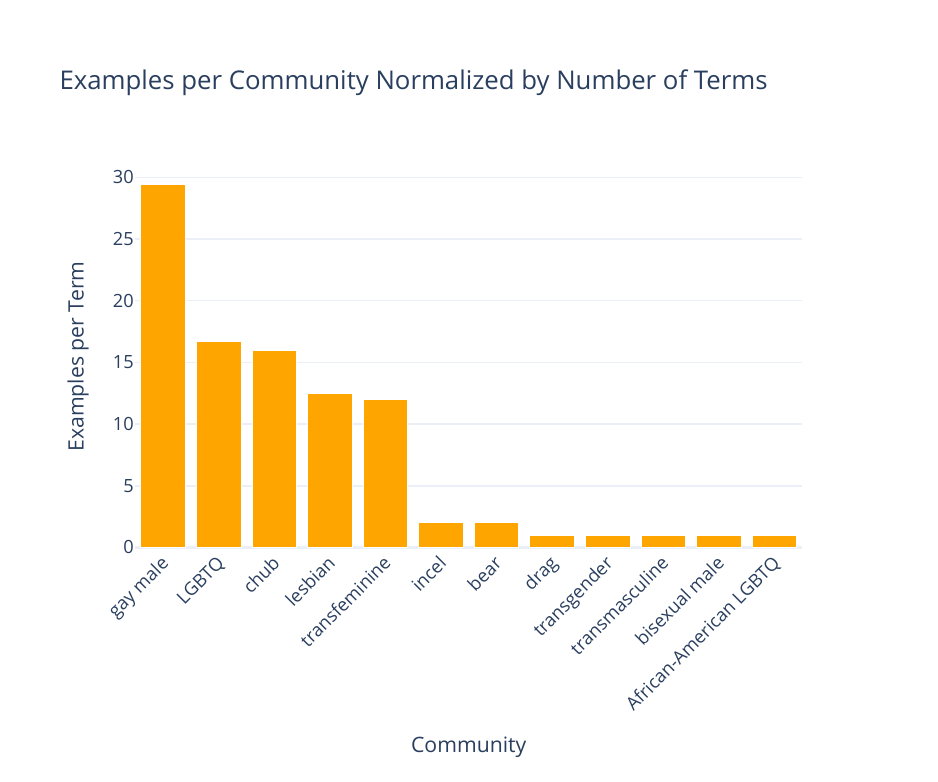}
        \label{fig:sub2}
    \end{subfigure}
    
    \vspace{0.5em}
    
    % Second row - 2 images
    \begin{subfigure}[b]{0.48\textwidth}
        \centering
        \includegraphics[width=\textwidth]{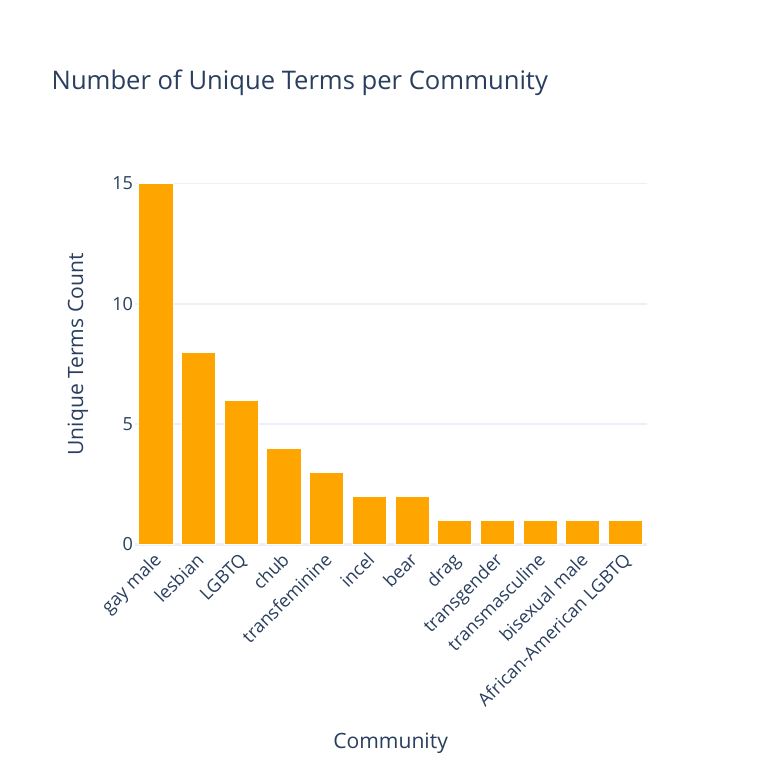}
        \label{fig:sub3}
    \end{subfigure}
    \hfill
    \begin{subfigure}[b]{0.48\textwidth}
        \centering
        \includegraphics[width=\textwidth]{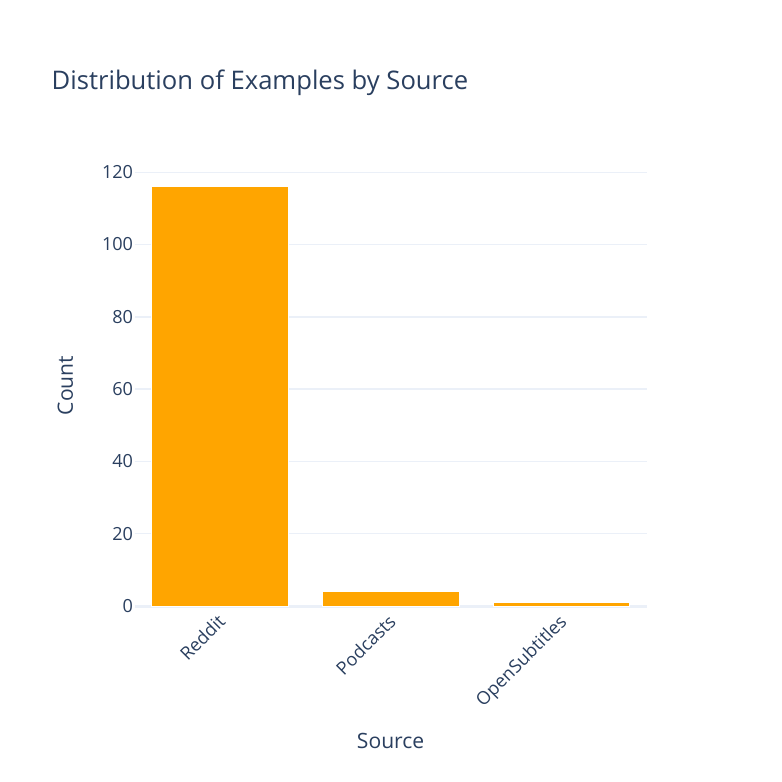}
        \label{fig:sub4}
    \end{subfigure}
    
    \vspace{0.5em}
    
    % Third row - 1 wide image
    \begin{subfigure}[b]{\textwidth}
        \centering
        \includegraphics[width=\textwidth]{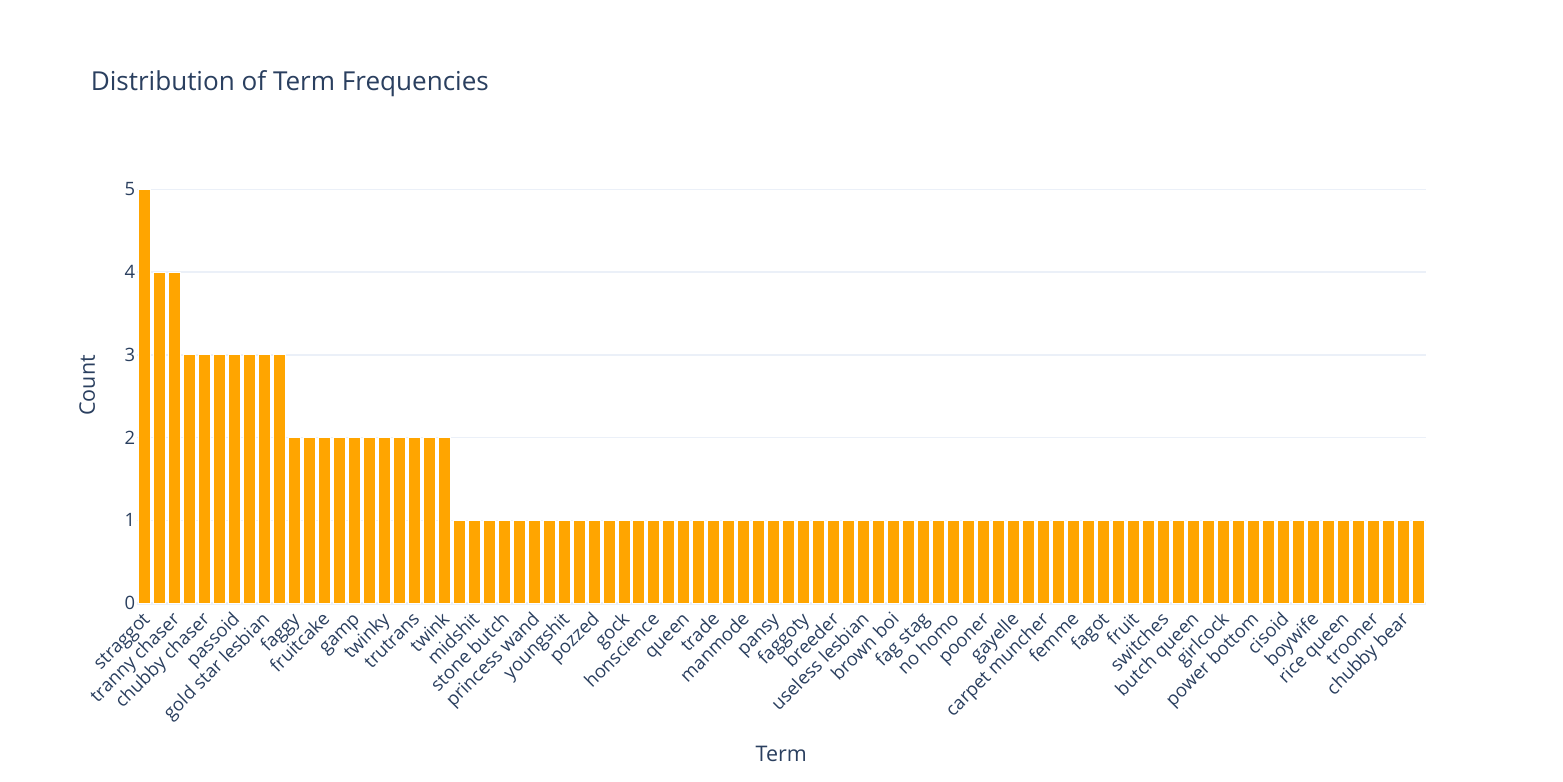}
        \label{fig:sub5}
    \end{subfigure}
    
    \caption{Distribution of terms, communities, and sources for the private version of \textsc{Slaying}.}
    \label{fig:priv_stats}
\end{figure}

\begin{figure}
  \centering
  \includegraphics[width=0.6\textwidth]{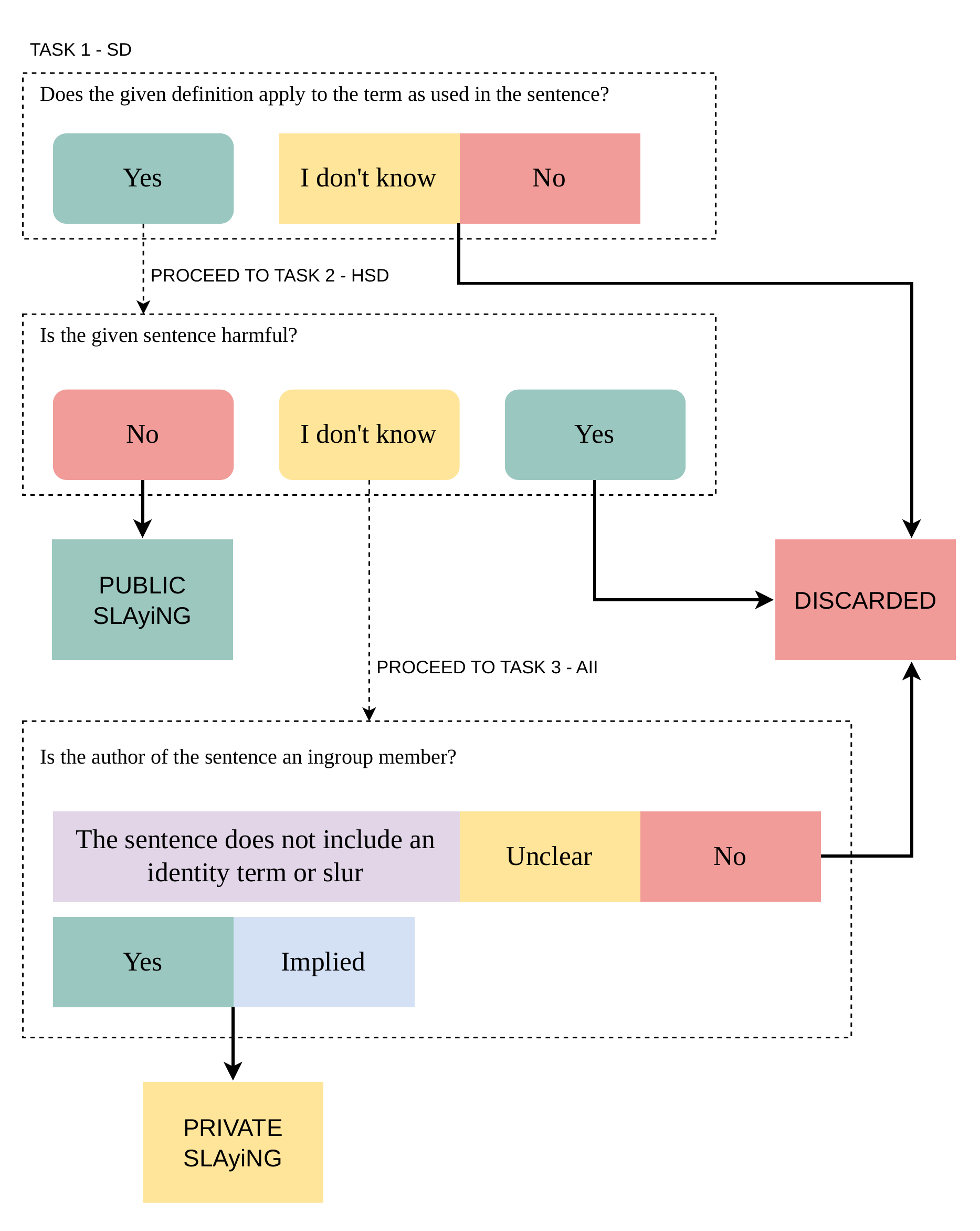}
  \caption{Each internally annotated example of \textsc{Slaying} was parsed though the depicted decision tree. Examples considered incorrect (\texttt{``No''} or \texttt{``I don't know''} for the sense disambiguation task) were completely discarded from the final dataset. This is also the case for harmful examples (\texttt{``Yes''} for the harmful speech detection task). Examples of uncertain harmfulness (\texttt{``I don't know''}) were included in a private version of \textsc{Slaying}, provided they were considered ingroup speech (author identity identification task).}
  \label{fig:decision_tree}
\end{figure}

\Cref{fig:decision_tree} depicts the decision tree for internal annotation of \textsc{Slaying}: only examples considered harmless were included in the public version of \textsc{Slaying} (3405 entries). The private version of \textsc{Slaying}, which contains an additional 121 entries, can be accessed through direct contact with the authors. These examples were considered potentially harmful, but ingroup speech of the queer community. \Cref{fig:public_stats} and \Cref{fig:priv_stats} depict statistics for the public and private versions of \textsc{Slaying}.

\subsection{Additional Questionnaire for Crowdsourced Annotators}
\label{app:Additional Questionnaire for Crowdsourced Annotators}
\begin{figure}
    \centering
    \includegraphics[width=0.8\textwidth]{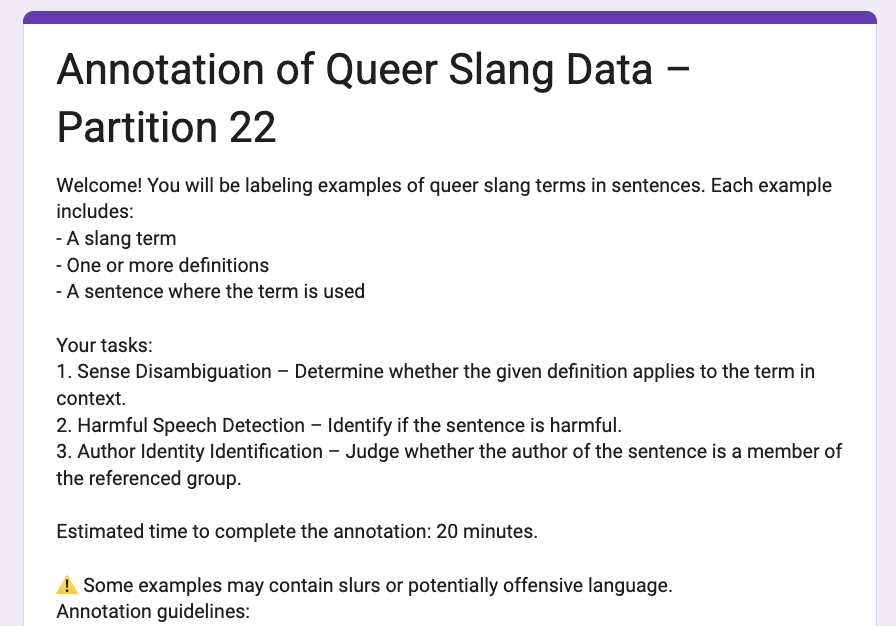}
    \caption{Task description at the start of the annotation google form for partition 22.}
    \label{fig:googleform}
\end{figure}

\begin{figure}
    \centering
    \includegraphics[width=0.4\textwidth]{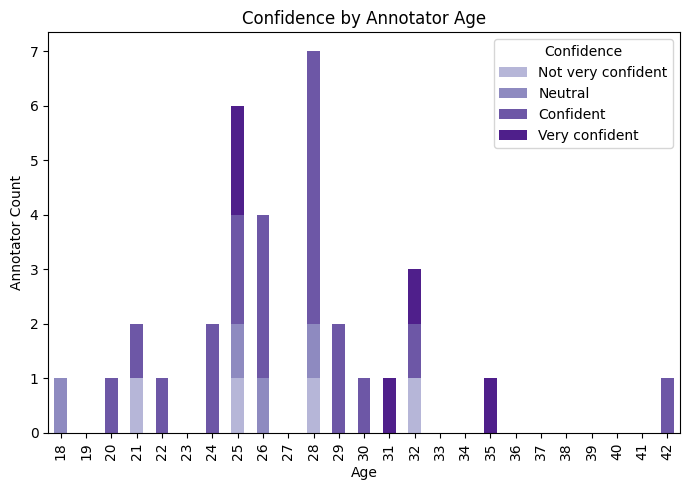}
    \caption{Annotator age distribution and according queer slang confidence.}
    \label{fig:age}
\end{figure}

\begin{figure}
        \centering 
        \includegraphics[width=\textwidth]{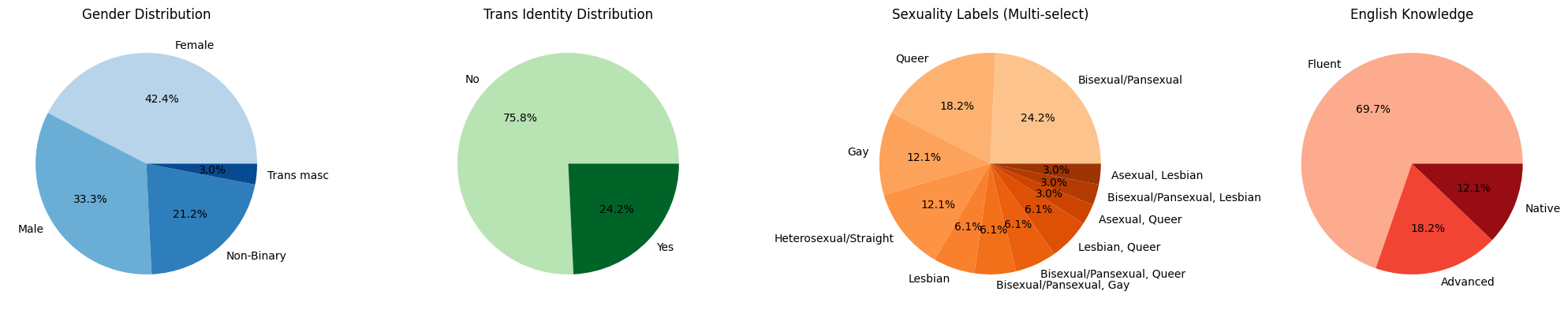}
   \caption{Demographic breakdown of annotators by gender, trans identity, sexuality (multi-label), and English knowledge.}
    \label{fig:demographic}
\end{figure}

\begin{figure}[t]
    \centering

    % -------- Top subfigure --------
    \begin{subfigure}[t]{\textwidth}
        \centering
        \includegraphics[width=\textwidth]{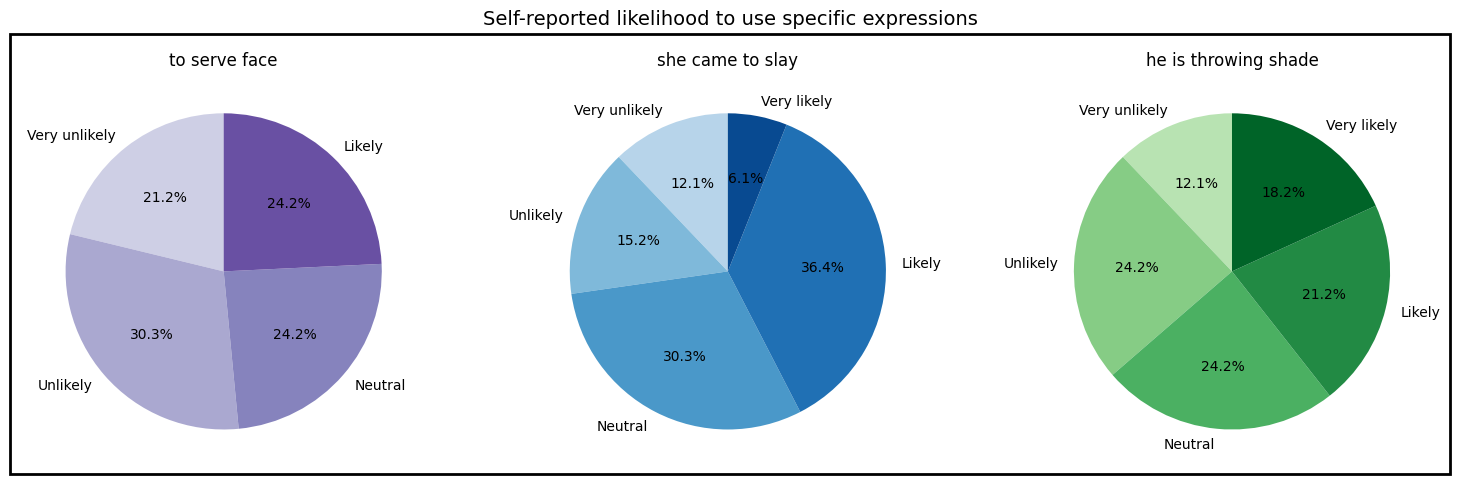}
        \caption{}
        \label{fig:screener_top}
    \end{subfigure}

    \vspace{0.5em}

    % -------- Bottom subfigure --------
    \begin{subfigure}[t]{\textwidth}
        \centering
        \includegraphics[width=\textwidth]{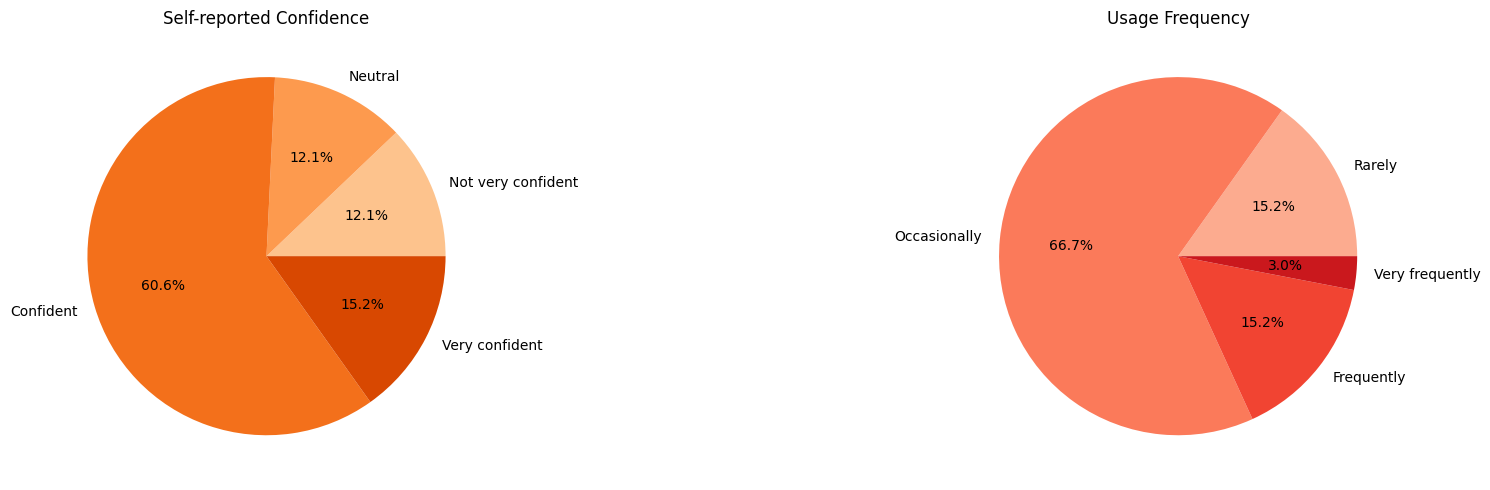}
        \caption{}
        \label{fig:screener_bottom}
    \end{subfigure}

    \caption{Annotator self-reported familiarity with English queer slang.}
    \label{fig:screener}
\end{figure}

%demographics, self-reported English knowledge and Queer Slang Screener
Participants registered voluntarily and were randomly assigned to one of 22 partitions, each corresponding to an online survey containing the respective data samples. During the registration process, participants provided their e-mail address where a google form shown in figure \ref{fig:googleform} was sent including the additional questionnaire, as well as data of a certain partition. Responses to the annotation forms were collected anonymously and all e-mail communication was deleted at the end of the annotation period. Filling out the annotation form, participants were asked to confirm that they had read the annotation guidelines established during internal annotation. Before starting annotation, they further completed a questionnaire covering demographic variables and self-reported familiarity with queer slang (\Cref{fig:questionnaire}). %We report descriptive statistics of these responses to the additional questionnaire below.
\begin{figure}[h]
\small
\begin{enumerate}
    \item \textbf{Have you read the annotation guidelines?} \\
     Yes, I have read the guidelines

    \item \textbf{How old are you?}   

    \item \textbf{What is your gender?} \\
    Male; Non-Binary; Female; Other (please specify)

    \item \textbf{Do you identify as trans?} \\
    Yes; No; Decline to State

    \item \textbf{What is your sexual orientation?}(Multiple Choice) \\
    Asexual; Bisexual/Pansexual; Gay; Heterosexual/Straight; Lesbian; Queer; Other (please specify)

    \item \textbf{How would you rate your English proficiency?} \\
    Native; Fluent; Advanced; Intermediate; Basic

    \item \textbf{We are trying to understand your experience with queer slang. Please look at these queer slang phrases. How likely are you to say or write something in the style of the phrases below?} 
    \begin{itemize}
        \item \textit{to serve face} \\
        Very likely;Likely; Neutral; Unlikely; Very unlikely
        \item \textit{she came to slay} \\
        Very likely;Likely; Neutral; Unlikely; Very unlikely
        \item \textit{he is throwing shade} \\
        Very likely;Likely; Neutral; Unlikely; Very unlikely
    \end{itemize}
    \item \textbf{How confident are you in your ability to understand queer slang?} \\
    Very confident;	Confident;	Neutral;	Not very confident;	Not at all confident
    \item \textbf{How frequently do you use queer slang?} \\
    Very frequently;	Frequently;	Occasionally;	Rarely;	Never

\end{enumerate}
\caption{Additional questionnaire for external participants.}
\label{fig:questionnaire}
\end{figure}
\paragraph{Demographic Information}
One annotator was excluded from annotation evaluation and demographic analysis due to low English proficiency and lack of familiarity with English queer slang. The remaining annotators span an age range from 18 to 42 (\Cref{fig:age}). The largest group of annotators identified as female, followed by 33\% male, 21\% non-binary, and one trans masculine annotator; overall, 24\% of annotators identified as trans. For sexuality, annotators were allowed to select multiple labels. The most common selections were bisexual/pansexual (24\%), queer (18\%), and gay (12\%). Regarding English proficiency, 12\% of annotators reported being native speakers and 70\% fluent speakers (\Cref{fig:demographic}).
\newpage
\paragraph{Queer Slang Familiarity}
As shown in \Cref{fig:screener_top}, the expressions \texttt{she came to slay} and \texttt{he is throwing shade} are more frequently reported as likely to be used by annotators, compared to \texttt{to serve face}, which only 24\% of annotators indicated they would likely use. Overall, 76\% of annotators report being confident or more in their understanding of queer slang, and 85\% report using such expressions at least occasionally (\Cref{fig:screener_bottom}). \Cref{fig:age} further illustrates the relationship between age and self-reported confidence, showing no clear or consistent pattern across age groups.
\label{App:Demographics}

\subsection{Inter-annotator Agreement}
\label{app:Inter-annotator Agreement}

\begin{figure}
   \centering
   \includegraphics[scale=0.45]{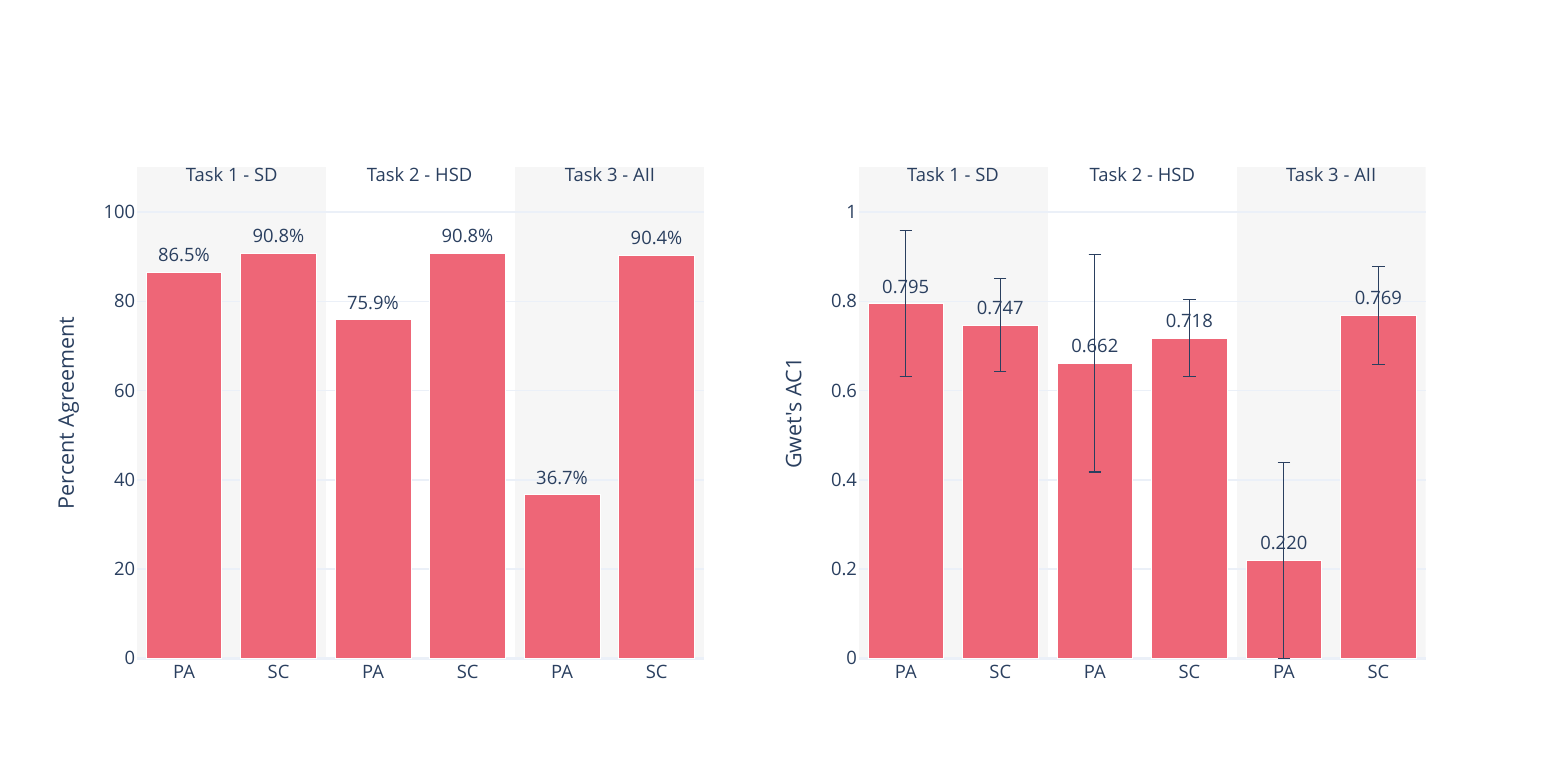}
   \caption{Inter-annotator agreement between internal and crowdsourced annotators, for the publicly available version of \textsc{Slaying}. Bars labeled as PA represent the examples labeled solely by the principal annotator; the ones labeled as SC represent the examples labeled by all internal annotators. \textbf{Agreement is moderate to high for sense disambiguation (Task 1) and harmful speech detection (Task 2), validating the correctness and safety of our data.}}
   \label{fig:crowdsouced_agreement_public}
 \end{figure}

\begin{figure}
  \centering
  \includegraphics[scale=0.45]{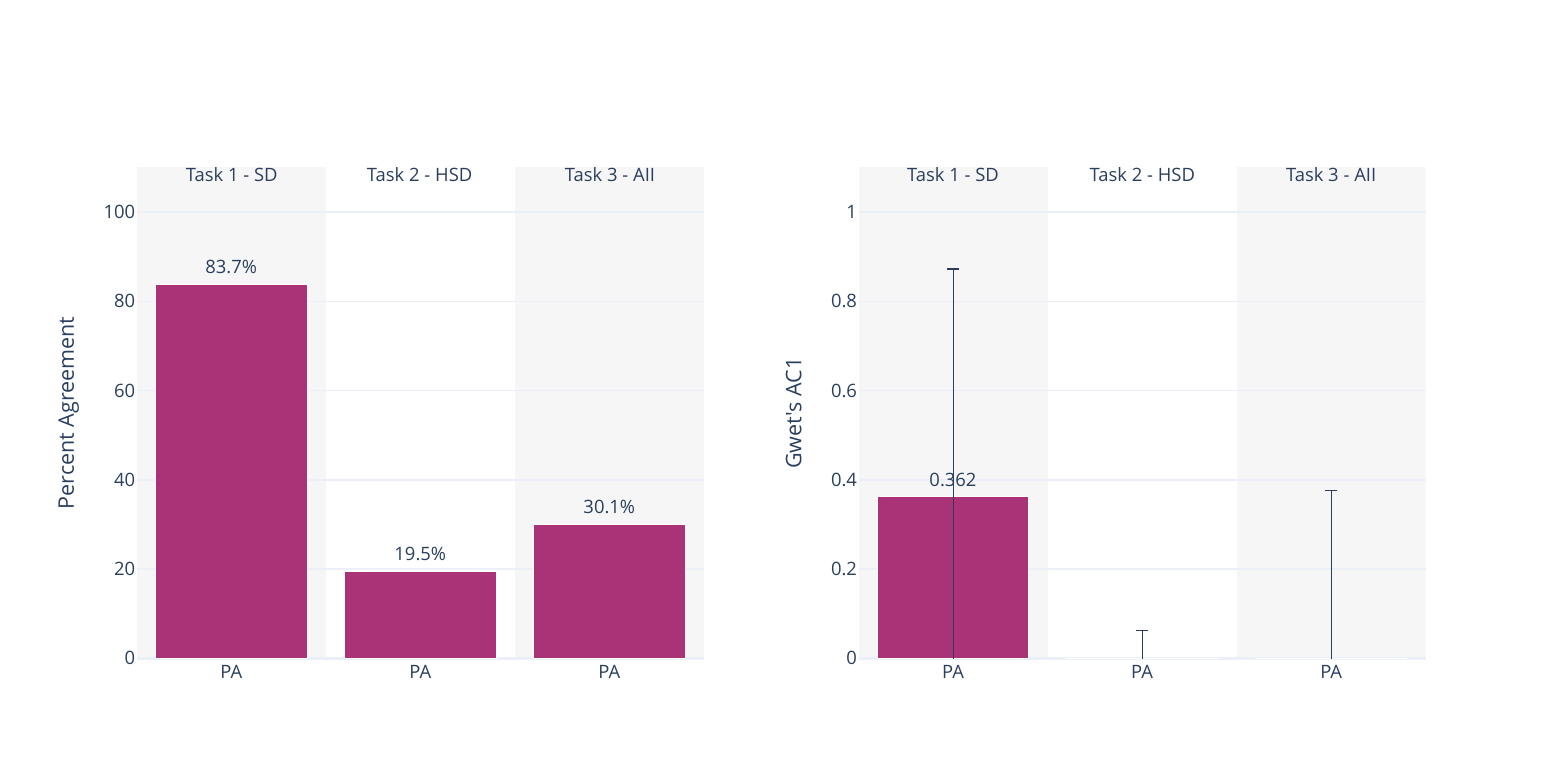}
  \caption{Inter-annotator agreement between internal and crowdsourced annotators, for the private version of \textsc{Slaying}. Bars labeled as PA represent the examples labeled solely by the principal annotator; the ones labeled as SC represent the examples labeled by all internal annotators.}
  \label{fig:crowdsouced_agreement_private}
\end{figure}

\begin{figure}
  \centering
  \includegraphics[scale=0.45]{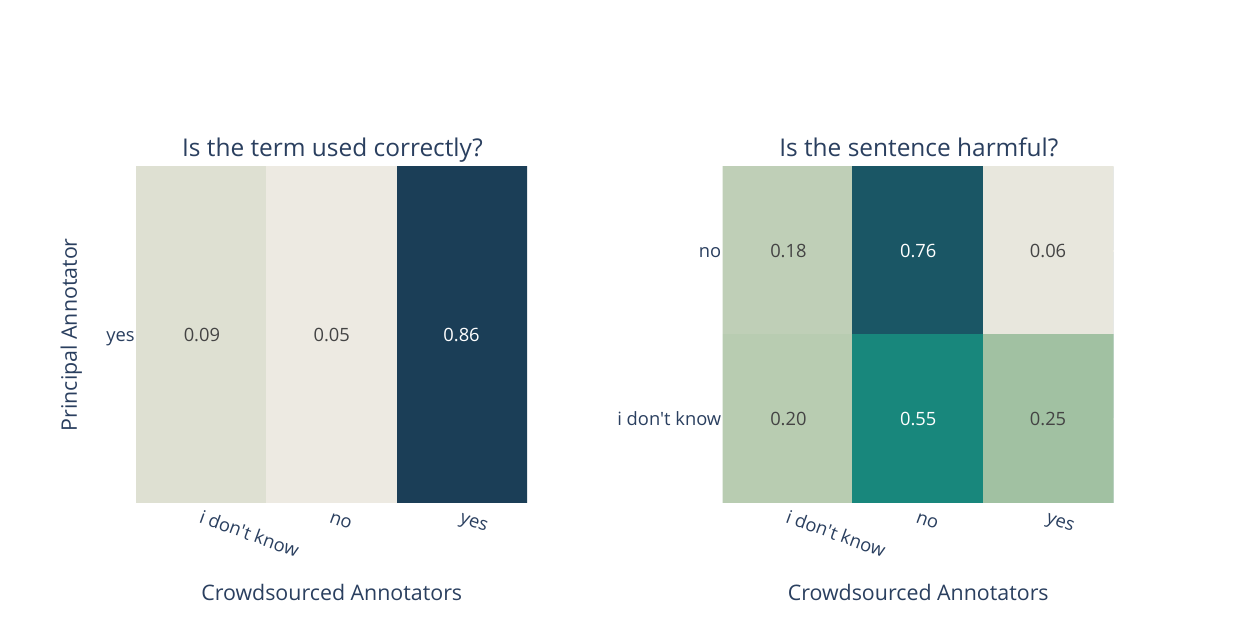}
  \caption{Label matrix for annotations of the PA and crowdsourced annotators. The majority of the private \textsc{Slaying} examples were labeled as not harmful by a crowdsourced annotator. Very few examples were labeled as not harmful by the PA and as harmful by the crowdsourced annotator (complete disagreement).}
  \label{fig:annotation_matrix}
\end{figure}

\begin{figure}
  \centering
  \includegraphics[scale=0.6]{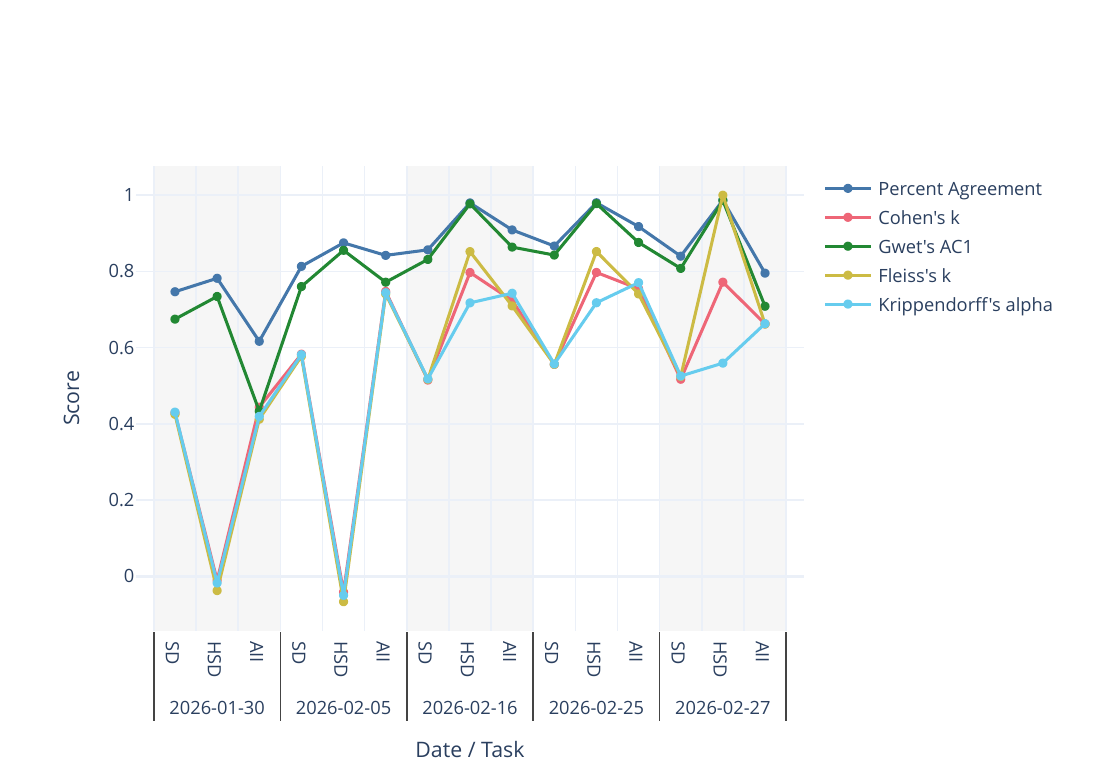}
  \caption{Agreement stability for the internal annotation process, including the pilot study and sanity check batches.}
  \label{fig:internal_stability}
\end{figure}

\begin{figure}
    \centering
    \includegraphics[scale=0.6]{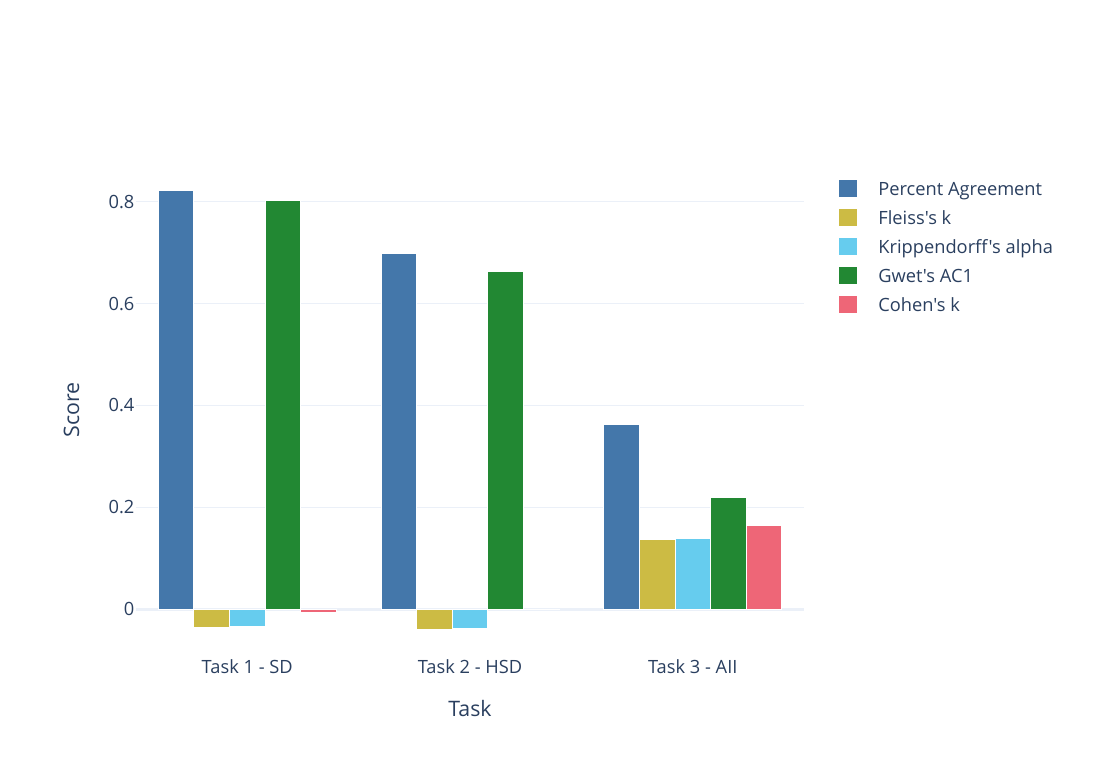}
    \caption{Inter-annotator variation: agreement metrics for the examples labeled by 2 crowdsourced annotators and the PA.}
    \label{fig:variation}
\end{figure}

This section contains extra results related to inter-annotator agreement, throughout the internal and crowdsourced annotation process. \Cref{fig:crowdsouced_agreement_public} and \Cref{fig:crowdsouced_agreement_private} depicts the agreement results for the public and private versions of \textsc{Slaying} -- here we only report percent agreement and Gwet's AC1 on the basis that Cohen's $\kappa$ is uninteresting ($0$ for tasks 1 and 2, which along with a high percent agreement indicates high data imbalance). \Cref{fig:annotation_matrix} depicts the distribution of labels as annotated by the PA and a crowdsourced volunteer. \Cref{fig:internal_stability} depicts the stability of the internal annotation process -- agreement between the principal annotator (PA) and the other internal annotators during the month-long effort. \Cref{fig:variation} depicts the inter-annotator agreement for the examples that were annotated by at least 2 crowdsourced annotators and the PA, effectively showing inter-annotator variation.

\subsection{Licensing}
\label{app:Licensing}

The licenses for the vocabularies sources of \textsc{Slaying} are as follows: GSSO falls under Apache License 2.0, lgbtDB under CC0, the Chew Glossary under CC BY-NC-SA 4.0, and Wiktionary under CC-BY-SA. Regarding sentence sources, the OpenSubtitles Corpus falls under ODC-BY. The short excerpts derived from podcast transcripts and Reddit remain the intellectual property of their respective creators and hosting platforms. The derived metadata and annotations produced in this work are released under the Creative Commons Attribution 4.0 International License (CC BY 4.0), where legally permissible. \textsc{Slaying} is intended for non-commercial academic research use only.

\section{Experimental Setup: Extra Results}
\label{app:Experimental Setup: Extra Results}

\subsection{Details of Case Study I}
\label{app:Details of Case Study I}

\begin{figure}[t]
  \centering
\includegraphics[width=\textwidth]{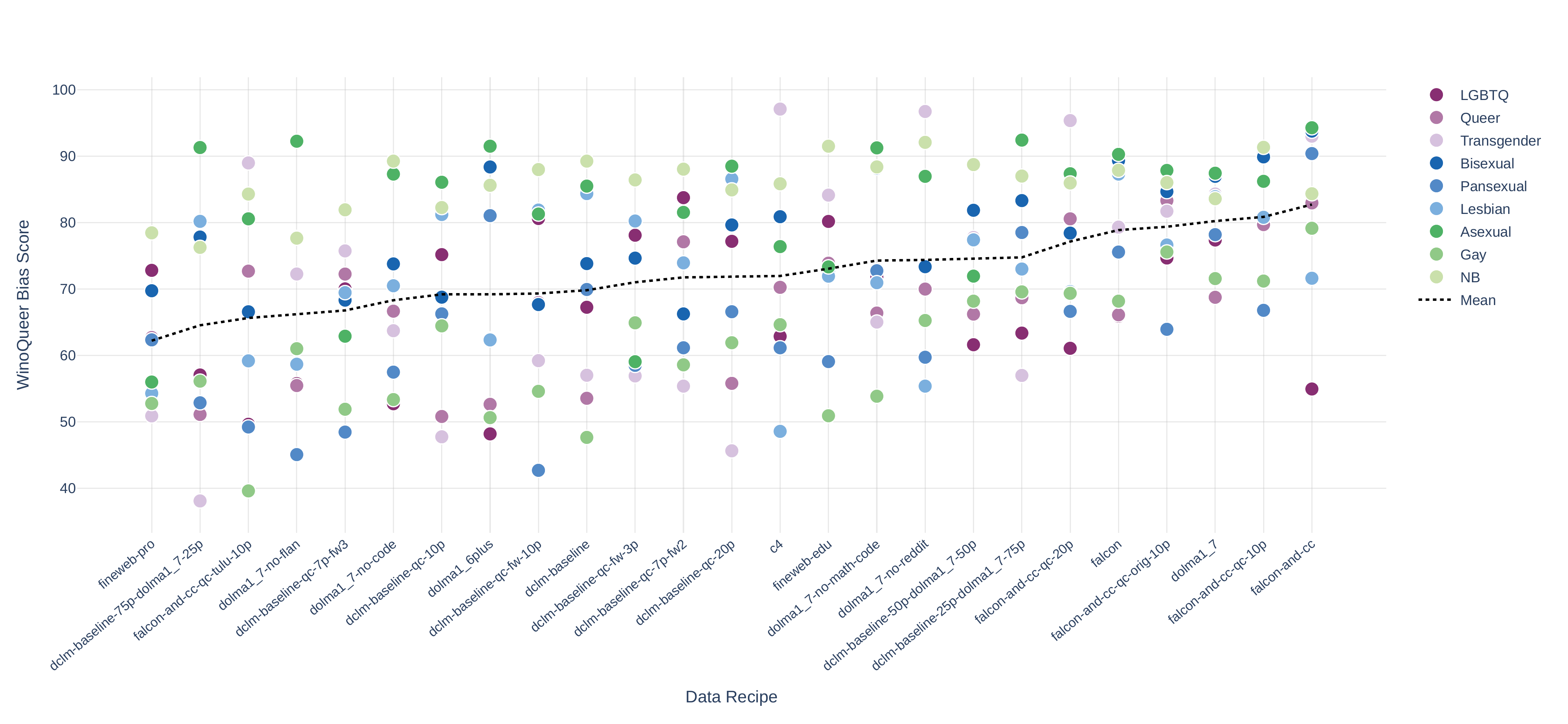}
  \caption{\textsc{WinoQueer} bias score per \textsc{Winoqueer} community, across all \textsc{DataDecide} data recipes. Recipes do affect which communities are targets of relative bias.}
  \label{fig:winoqueer_communities}
\end{figure}

\paragraph{Perplexity} Following \citet{gao2020pile, magnusson2024paloma}, we use BPB as a proxy for language fit. Bits per byte is preferred over bits per
character or perplexity due to its invariance to different tokenization schemes and the ambiguity of measuring characters in Unicode. BPB normalizes the log likelihood $\ell$ over documents by the count of UTF-8 encoded bytes, $B$: 

$$\textbf{BPB} = \frac{1}{B}\log_2(e^{-\ell}) = \frac{-\ell}{B\ln(2)}$$ 

\begin{figure}[t]
  \centering
\includegraphics[width=0.6\textwidth]{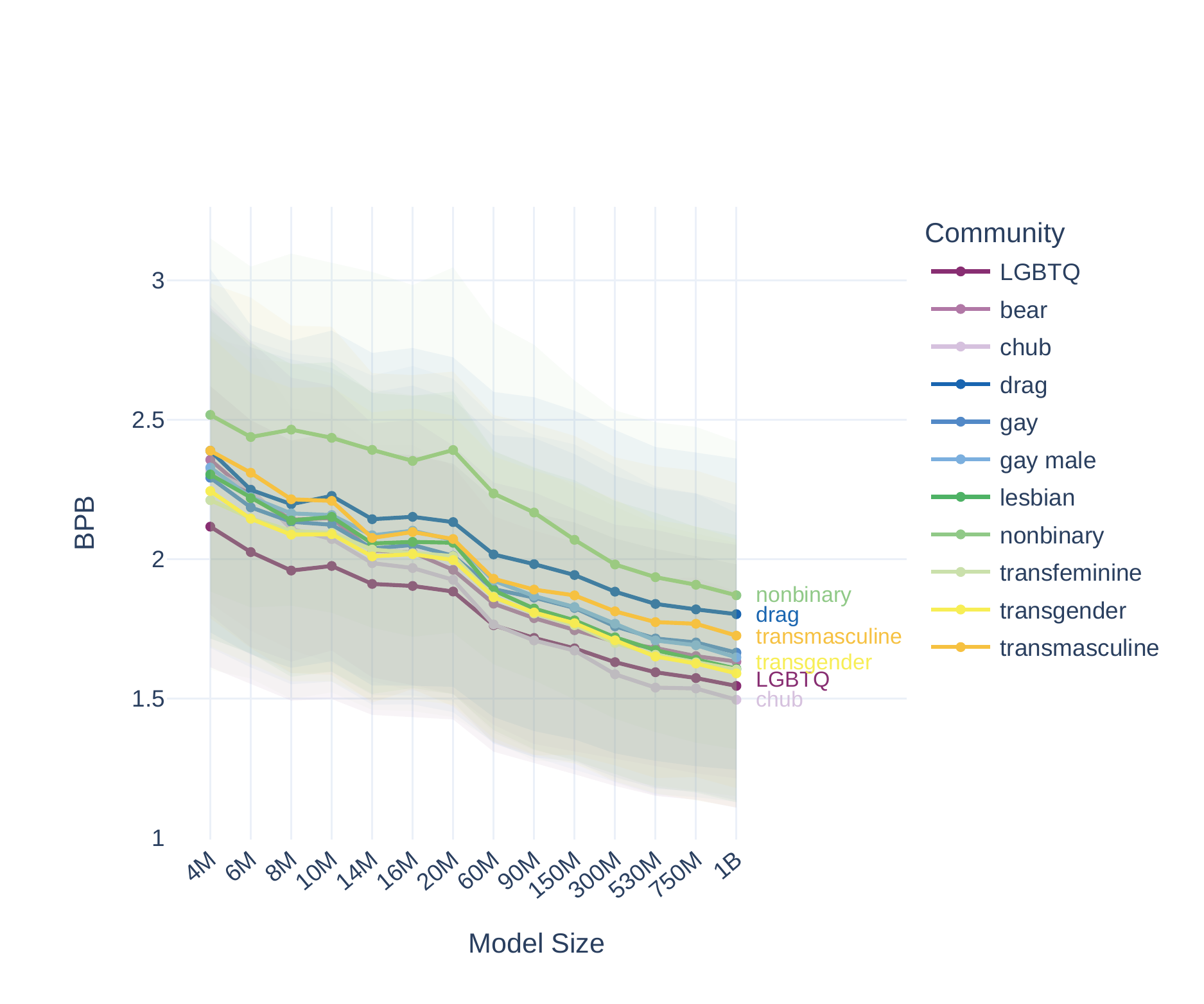}
  \caption{BPB across model size of the \textsc{DataDecide} models trained on \texttt{dolma1\_7}, for the standardized versions of \textsc{Slaying} (left) and for each queer subcommunity of \textsc{Slaying} (right). Performance on the original version of \textsc{Slaying} is worse relative to its standard English versions. Some queer subcommunities see less diminishing returns than others -- for example, performance on nonbinary slang inches closer to other subcommunities as model size increases.}
  \label{fig:model_size}
\end{figure}

\paragraph{Impact of Model Size on Language Fit to Queer Slang} We employ the full breadth of model sizes of \textsc{DataDecide} trained on \texttt{dolma1\_7}, ranging from 4M to 1B parameters. \Cref{fig:model_size} depicts results for each queer subcommunity present in \textsc{Slaying}, confirming a performance bias against nonbinary and drag slang. All communities show a similar trend of performance gain as model size increases.

\paragraph{Hyperparameters and Details on Continued Pretrained Olmo} We train a 1B OLMo\footnote{\url{https://huggingface.co/allenai/OLMo-1B-hf}} for 1 epoch, with a batch size of 8, weight decay of 0.01, and a learning rate of 5e-06 (AdamW optimizer). We used a random 80/20 train-test split for \textsc{Slaying}. We performed a small hyperparameter search on 5 learning rate values ($lr = [1e-6, 3e-6, 5e-6, 1e-5, 2e-5]$), while keeping the rest of the parameters (due to compute and time restrictions). We chose the best model (regarding improvement on \textsc{Slaying} and \textsc{WinoQueer}) to depict in the main document (\Cref{fig:olmo_improvement}). Results for all models are depicted in \Cref{tab:hyperparameter_results}

\begin{table}[]
\centering
\caption{OLMo-1B performance across different learning rate configurations.}
\begin{tabular}{lccccc}
\hline
\textbf{Metric} & \textbf{Run 0} & \textbf{Run 1} & \textbf{Run 2} & \textbf{Run 3} & \textbf{Run 4} \\
 & \textbf{lr=1e-6} & \textbf{lr=3e-6} & \textbf{lr=5e-6} & \textbf{lr=1e-5} & \textbf{lr=2e-5} \\
\hline
Base BpB & 1.6102 & 1.6102 & 1.6102 & 1.6102 & 1.6102 \\
Finetuned BpB & 1.3746 & 1.4269 & 1.3408 & 1.3574 & 1.3948 \\
BpB Improvement & 0.2355 & 0.1833 & 0.2694 & 0.2528 & 0.2154 \\
\hline
\multicolumn{6}{c}{\textit{WinoQueer Accuracy (\%)}} \\
\hline
Overall & 54.88 & 53.60 & 50.99 & 47.54 & 36.69 \\
LGBTQ & 40.48 & 38.26 & 38.02 & 28.69 & 10.81 \\
Queer & 51.45 & 59.44 & 53.76 & 59.22 & 46.04 \\
Transgender & 42.39 & 49.06 & 47.41 & 55.64 & 47.74 \\
Bisexual & 81.33 & 75.18 & 71.25 & 62.45 & 56.17 \\
Pansexual & 63.03 & 56.21 & 54.86 & 56.00 & 56.00 \\
Lesbian & 53.56 & 50.26 & 37.15 & 47.32 & 12.95 \\
Asexual & 90.79 & 90.48 & 86.14 & 86.97 & 80.04 \\
Gay & 33.15 & 21.64 & 22.59 & 1.05 & 1.22 \\
NB & 89.20 & 85.85 & 87.59 & 95.15 & 88.22 \\
\hline
\end{tabular}
\label{tab:hyperparameter_results}
\end{table}

\subsection{Details of Case Study II}
\label{app:Details of Case Study II}

This section contains a more direct comparison with the work of \citet{dorn2024harmful} on toxicity classifier performance on sentences containing reclaimed queer slurs. The \textsc{Slaying} public dataset, as is, can only be used to calculate a false positive rate in toxicity/harmful speech classification, as all examples were manually labeled as harmless. As a byproduct of our extensive annotation process, we create a small evaluation benchmark for harmful speech detection, with examples from the raw version of \textsc{Slaying}: this benchmark contains 790 examples (395 labeled as \texttt{Yes} for the task of harmful speech detection, 395 labeled as \texttt{No}, with a majority label), covering 316 slang terms. We test the \textsc{Dolma} FastText hate-speech classifier (part of its content filtering pipeline), trained on Jigsaw Toxic Comments \citep{soldaini2024dolma, jigsaw-toxic-comment-classification-challenge}; the original \textsc{Detoxify} model, for its usage in computational social science and queer NLP works \citep{dorn2024non, dorn2024harmful}; and the two \textsc{ToxiGen} models (HateBERT and Roberta), due to their high performance on implicit hate-speech detection tasks \citep{hartvigsen2022toxigen}. Results are depicted in \Cref{tab:classifier_metrics}. As \citet{dorn2024harmful} also use Detoxify in their classifier suite, we can do a direct comparison -- they achieve 0.25 F1 and 0.59 F1 for ingroup speech and outgroup speech, respectively. Detoxify performs better on \textsc{Slaying}, confirming that examples containing reclaimed slurs are more prone to being wrongly classified by these models. However, as discussed in the main document (section \Cref{subsec: Case Study II}), other terms and features of queer language can also be misclassified as toxic. In \textsc{Slaying} the best performing model is \textsc{ToxiGen}'s HateBERT.

\begin{table}[]
\centering
\caption{Classifier Performance (threshold $> 0.5$)}
\label{tab:classifier_metrics}
\begin{tabular}{lccc}
\toprule
Classifier & Precision & Recall & F1 \\
\midrule
Dolma & 0.8227 & 0.2937 & 0.4328 \\
Detoxify • Original & 0.8037 & 0.5494 & 0.6526 \\
ToxiGen • HateBERT & 0.7231 & 0.8000 & 0.7596 \\
ToxiGen • RoBERTa & 0.8182 & 0.5468 & 0.6555 \\
\bottomrule
\end{tabular}
\end{table}

\subsection{How Queer Slang gets Lost in the Data Pipeline: The Case of \textsc{Dolma}}

\begin{figure}
  \centering
  \includegraphics[scale=0.375]{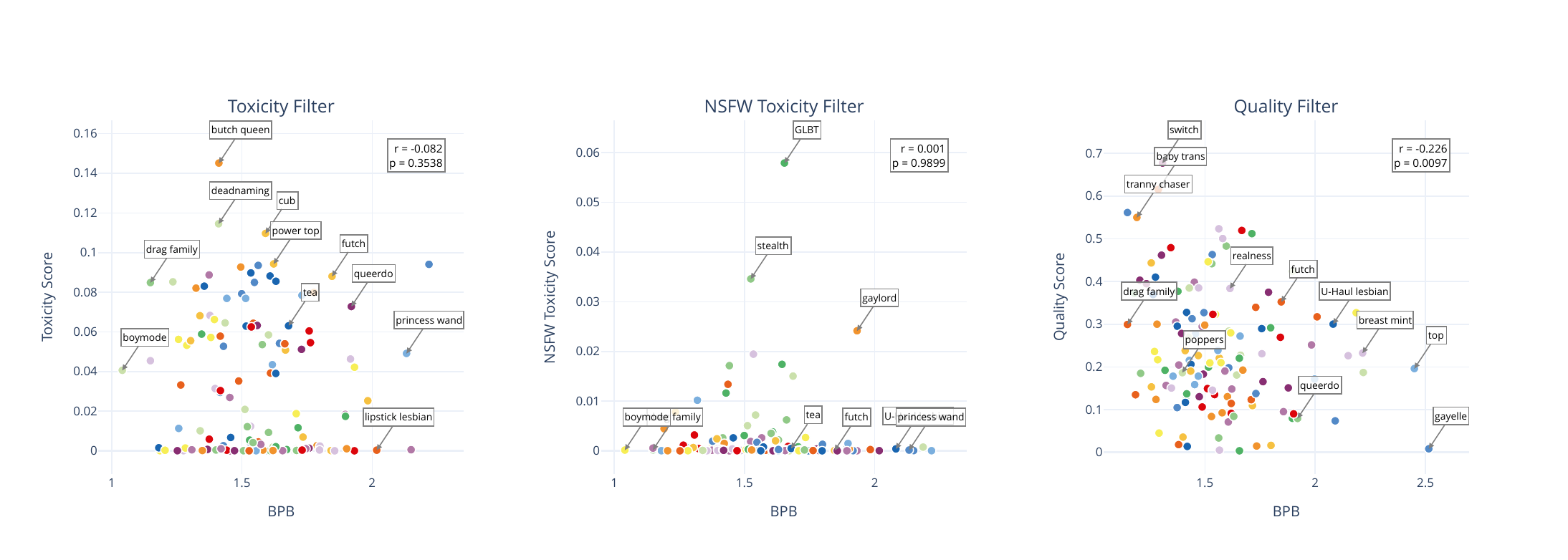}
  \caption{Term-wise comparison of BPB on the \textsc{DataDecide} 1B model trained on \texttt{dolma-1.7}, and quality and toxicity scores of the \textsc{Dolma} pipeline content filters. Toxicity does not significantly correlate with BPB, but quality does, negatively so -- ``low-quality'' examples are filtered out of the data pipeline, leading to a lower BPB on the trained model.}
  \label{fig:dolma_analysis}
\end{figure}

In this section we make the connection between the two metrics for linguistic bias we explore in this work: language model fit (perplexity, or BPB) and toxicity. Toxicity classifiers, when used as a component of the pretraining data pipeline, can directly influence the fit of a language model on minority or rare vernacular. For instance, if a certain slang term is constantly mislabeled as toxic, it is left out of the pretraining corpus, and therefore unable to be accurately represented and processed by the resultant model. To analyze this hypothetical phenomenon, we leverage artifacts of the \textsc{Dolma} 1.7 \citet{soldaini2024dolma} corpus: its quality and content (toxic hate-speech and toxic NSFW content) filters, and the \textsc{DataDecide} 1B Olmo trained on \textsc{Dolma} 1.7 (same model family as used in \Cref{sec:Experimental Setup}). We calculate the quality (higher quality score, higher quality sentence) and toxicity (higher toxicity score, more toxic sentence) of \textsc{Slaying}, per term -- only including terms with at least 10 instances. Results are depicted in \Cref{fig:dolma_analysis}. Toxicity does not significantly correlate with BPB, but quality does, negatively so -- ``low-quality'' examples are filtered out of the data pipeline, leading to a lower BPB on the trained model. This suggests that toxicity classifiers are not the culprit behind lack of representation of queer slang in data, but quality filters might be. We encourage future work in this direction. %todo weird qc results (big variance. etc) to show why we didnt go in this direction, and say that we should test it on bigger docs

\section{Compute Resources}
\label{app:Compute Resources}

For all present experiments, we used a NVIDIA RTX A6000 and a NVIDIA A100-SXM4-80GB GPU. Continued pretraning of the OLMo models described in \Cref{subsec: Case Study II} and \Cref{app:Details of Case Study II} takes under 2h on a  NVIDIA A100-SXM4-80GB. Replication of all experiments in this paper takes under 24h (approximately) under the current setup, the largest bottleneck being testing all \textsc{DataDecide} recipes on the \textsc{WinoQueer} benchmark. AI assistants (code LLMs) were used to improve the aesthetic and readability of our visualizations and tables, and for bash scripting (these scripts were used in the experiments of \Cref{sec:Experimental Setup}).

\end{document}